\documentclass{article} 
\usepackage{iclr2027_conference,times}

\usepackage{amsmath,amsfonts,bm}

\def\eqref#1{equation~\ref{#1}}

\def\1{\bm{1}}

\DeclareMathAlphabet{\mathsfit}{\encodingdefault}{\sfdefault}{m}{sl}
\SetMathAlphabet{\mathsfit}{bold}{\encodingdefault}{\sfdefault}{bx}{n}

\usepackage{hyperref}
\usepackage{url}
\usepackage{graphicx}
\usepackage{booktabs}
\usepackage{multirow}
\usepackage{comment}

\title{Have You Seen It Clearly? Uncertainty-Guided Belief Refinement for Zero-Shot Object Navigation}

\author{
\textbf{Jingzhi Huang}\textsuperscript{1} \hspace{1.5em}
\textbf{Junkai Huang}\textsuperscript{2} \hspace{1.5em}
\textbf{Yuhang Peng}\textsuperscript{3} \hspace{1.5em}
\textbf{Keyu Mu}\textsuperscript{4} \hspace{1.5em}
\textbf{Jiaru Zhong}\textsuperscript{1} \and
\textbf{Yi Wang}\textsuperscript{1} \hspace{1.5em}
\textbf{Lujia Wang}\textsuperscript{4} \hspace{1.5em}
\textbf{Lap-Pui Chau}\textsuperscript{1} \hspace{1.5em}
\textbf{Siyang Gao}\textsuperscript{5}\textsuperscript{\dag} \\[1.5ex]
\textsuperscript{1}Hong Kong Polytechnic University \quad
\textsuperscript{2}Institute of Automation, Chinese Academy of Sciences \\
\textsuperscript{3}DeepRobotics \quad
\textsuperscript{4}Hong Kong University of Science and Technology (Guangzhou) \\
\textsuperscript{5}City University of Hong Kong \\[1ex]
\texttt{jingzhi.huang@connect.polyu.hk}, \texttt{siyangao@cityu.edu.hk} \\
\textsuperscript{\dag}Corresponding author \\
\url{https://github.com/kim-os/RefBeliefNav}
}

\iclrfinalcopy 
\begin{document}

\maketitle
\fancyhead{} 
\renewcommand{\headrulewidth}{0pt} 
\fancyfoot[C]{\thepage} 

\begin{abstract}
Zero-Shot Object Navigation (ZSON) requires an agent to locate target objects in unfamiliar environments without additional training for the test categories or environments. Recent approaches leverage large language models and vision-language models to construct spatial representations and guide exploration. However, whether the available evidence sufficiently supports a navigation choice, and how new evidence changes the search value of previously visited locations, remain insufficiently addressed. We propose RefBeliefMap, an uncertainty-guided mapping framework that distinguishes target-location belief from spatial and semantic evidence uncertainty. By jointly modeling intra-observation uncertainty and inter-observation disagreement over a sliding memory window, RefBeliefMap assesses the reliability of both evidence channels. When competing frontiers remain insufficiently resolved, these uncertainty estimates guide active observation and semantic reasoning to refine the evidence supporting target-location beliefs before navigation commitment. Building on the refined map, we introduce RefBeliefNav, which jointly evaluates unexplored frontiers and historical waypoints according to their target-finding potential, expected information gain, and navigation cost. This unified objective allows historical locations to be reconsidered as new evidence changes their search value, coordinating exploration and revisitation. Experiments on HM3D and MP3D demonstrate state-of-the-art performance among the evaluated zero-shot baselines, with ablation studies examining the contributions of modules. Real-world experiments on a Unitree Go2 quadruped robot further demonstrate the sim-to-real transferability of our method.
\end{abstract}

\section{Introduction}

Zero-shot object navigation (ZSON) requires an embodied agent to locate and reach a target specified by a category name or natural-language description in an unfamiliar environment, without additional training for the test categories or environments. This capability supports applications such as household assistance and autonomous inspection \cite{yokoyama2024vlfm,kuang2024openfmnav,cai2024pixnav,ginting2024seek}, while open-vocabulary benchmarks extend navigation beyond predefined target categories \cite{yokoyama2024hm3dovon}. By transferring prior knowledge to new goals and environments, ZSON reduces the need for deployment-specific data collection and model adaptation, offering a scalable approach to navigation across diverse settings.

Recent advances in large language models (LLMs) and vision-language models (VLMs) have enabled ZSON methods to leverage transferable semantic knowledge for spatial search. Many modular approaches comprise two recurring stages: map construction and navigation decision-making. For map construction, VLFM, OpenFMNav, and GAMap encode semantic relevance and affordance cues in two-dimensional maps \cite{yokoyama2024vlfm,kuang2024openfmnav,cai2024pixnav,yuan2024gamap}, while BeliefMapNav integrates semantic priors and observations into a three-dimensional voxel-based belief map \cite{zhou2025beliefmapnav}. Complementary representations include open-vocabulary and hierarchical scene graphs \cite{gu2024conceptgraphs,werby2024hovsg,yin2024sgnav,yin2025unigoal} and multi-scale Gaussian-language maps \cite{zhang2026glmap}. For decision-making, strategies range from greedy frontier selection to topology-aware and probabilistic planning, coarse-to-fine localization, and adaptive exploration \cite{wu2024voronav,ginting2024seek,zhang2025fbn,zhang2024trihelper,cao2025cognav,zhang2025apexnav,nie2025wmnav}. Recent methods further incorporate benefit-aware reasoning and retrieval of past navigation experience \cite{chen2026chainofsearch,wang2026trajrag}. Despite these advances, two limitations remain insufficiently addressed. First, most existing methods directly rank candidate navigation goals using currently available observations and semantic cues, but rarely assess whether the available evidence is itself sufficient to reliably distinguish among competing choices.  Second, although recent methods increasingly preserve or retrieve historical navigation information, they generally do not explicitly re-evaluate the search value of previously visited locations as new evidence changes the agent's belief about the environment. As a result, locations that appeared uninformative when first encountered may remain neglected even after subsequent observations make them relevant to the target search. These limitations motivate explicit assessment of evidence sufficiency during exploration and evidence-driven re-evaluation of historical locations.

We introduce RefBeliefMap, an uncertainty-guided mapping framework that separately represents target-location belief and spatial and semantic evidence uncertainty. A sliding observation window supports temporal evidence assessment, while spatial and semantic uncertainty guide active observation and semantic reasoning, respectively. These complementary mechanisms improve observational coverage and resolve ambiguous interpretations to refine the belief map. Building on this representation, RefBeliefNav jointly evaluates unexplored frontiers and historical waypoints according to their target-finding potential, expected information gain, and navigation cost. By updating candidate values as new evidence becomes available, the planner coordinates exploration and revisitation under a unified objective.

Our contributions are threefold: 1) We introduce RefBeliefMap, which distinguishes target-location
belief from spatial and semantic evidence uncertainty, enabling
uncertainty-guided active observation and semantic reasoning
to refine evidence for navigation decisions.
2) We propose RefBeliefNav, which unifies exploration and revisitation through belief-dependent evaluation of unexplored frontiers and historical waypoints.
3) We validate RefBeliefNav through simulation and real-world
experiments, achieving 69.2\% SR and 35.3\% SPL on HM3D
and 49.1\% SR and 18.5\% SPL on MP3D, outperforming the
evaluated zero-shot baselines on both metrics. Deployment
on a Unitree Go2 quadruped robot further demonstrates
its sim-to-real transferability.

\section{related work}
\subsection{Object Navigation}
Object navigation aims to guide a robot to locate a specified target object in an unknown environment. Existing approaches broadly fall into two groups. Training-based methods, including reinforcement learning and imitation learning, typically require extensive training on task-specific datasets \cite{ramakrishnan2022poni-train1,train2,train3,train4,train5}. Zero-shot methods instead exploit knowledge from pre-trained vision-language models (VLMs) \cite{Blip-2,Clip} and large language models (LLMs) to support navigation in unseen environments \cite{kuang2024openfmnav,cao2025cognav}. Following this paradigm, our framework uses LLM-derived semantic priors to initialize target-location beliefs at the start of a task and invokes online reasoning when semantic uncertainty is high to reassess ambiguous cues and inform belief refinement.

\subsection{Uncertainty-Aware Object Navigation}
Recent object navigation methods increasingly account for the reliability of semantic maps and accumulated observations. CARe estimates uncertainty through prediction confidence and multi-view consistency to revise decisions based on pre-explored semantic maps \cite{ko2025care}. OneMap uses probabilistic semantic updates to mitigate feature-extraction errors and guide exploration, while BeliefMapNav integrates semantic priors and observations into voxel-level target beliefs \cite{busch2025onemap,zhou2025beliefmapnav}. Extending this perspective to changing environments, IGV-RRT combines uncertainty-aware scene priors with confidence-weighted online semantic observations for object search \cite{zhang2026igvrrt}. Recent evidence-driven approaches further connect observation reliability with navigation: AECNav consolidates target and negative evidence into cluster-level log-odds beliefs and actively acquires informative observations, whereas AeroBelief separates contextual plausibility from qualified target-specific evidence for aerial ObjectNav \cite{liu2026aecnav,xiao2026aerobelief}. These advances establish the importance of evidence accumulation, reliability assessment, and active search. Although these methods assess evidence reliability, they do not fully characterize uncertainty in continuous robot navigation. At each step, uncertainty may arise from insufficient spatial observations or ambiguous semantic cues, while across steps, inconsistencies between current and previous observations introduce an additional source of uncertainty.

\subsection{Exploration Strategies for Object Navigation}
Exploration strategies for object navigation use geometric and semantic information to prioritize search locations. PONI learns complementary potential functions to guide search, while Semantic Frontiers and ESC incorporate semantic context and commonsense constraints into frontier selection \cite{ramakrishnan2022poni-train1,chen2023semanticfrontiers,zhou2023esc}. VoroNav uses topological paths to support waypoint reasoning, and SEEK formulates probabilistic search using relational semantic knowledge \cite{wu2024voronav,ginting2024seek}. More recent approaches adapt the search process to available evidence. ApexNav switches between semantic and geometric exploration according to the distribution of semantic cues, while BeliefMapNav optimizes frontier visitation sequences using observation beliefs and travel costs \cite{zhang2025apexnav,zhou2025beliefmapnav}. Historical observations also support further search: SSMG-Nav maintains persistent semantic memory and revisits mapped locations to recover potential missed detections after exploration is complete \cite{niu2026ssmgnav}.However, frontier sequencing and staged revisiting do not directly establish when revisiting is preferable to further exploration. RefBeliefNav evaluates historical waypoints alongside unexplored frontiers using updated beliefs, expected information gain, and navigation cost, dynamically determining which waypoint to revisit.

\section{Method}
\subsection{Task Definition}
We formulate the ZSON task as follows. The robot is required to locate a specified target object in an unfamiliar environment that has not been encountered previously. The target object \(c\) is specified by a free-form natural-language description. At each discrete time step \(t\), the agent observes the environment through an RGB-D sensor while estimating its relative state \(s_t=(x_t,r_t)\) using on-board odometry, where \(x_t\in\mathbb{R}^3\) denotes the agent position and \(r_t\in SO(3)\) represents its orientation. Based on the current observation and state estimate, the agent selects a discrete action \(a_t\) from the action space: $\mathcal{A}$: = \{\texttt{MOVE\_FORWARD}, \texttt{TURN\_LEFT}, \texttt{TURN\_RIGHT}, \texttt{LOOK\_UP}, \texttt{LOOK\_DOWN}, \texttt{STOP}\}. Motion is executed using fixed action increments, \(0.25\,\mathrm{m}\) for forward translation and \(30^\circ\) for rotational actions. An episode is considered successful if the agent issues the \(\texttt{STOP}\) action within a distance threshold of \(0.1\,\mathrm{m}\) from the target object before reaching the maximum interaction budget of 500 steps.

\subsection{System overview}

\begin{figure*}[t]
    \centering
    \includegraphics[width=\linewidth]{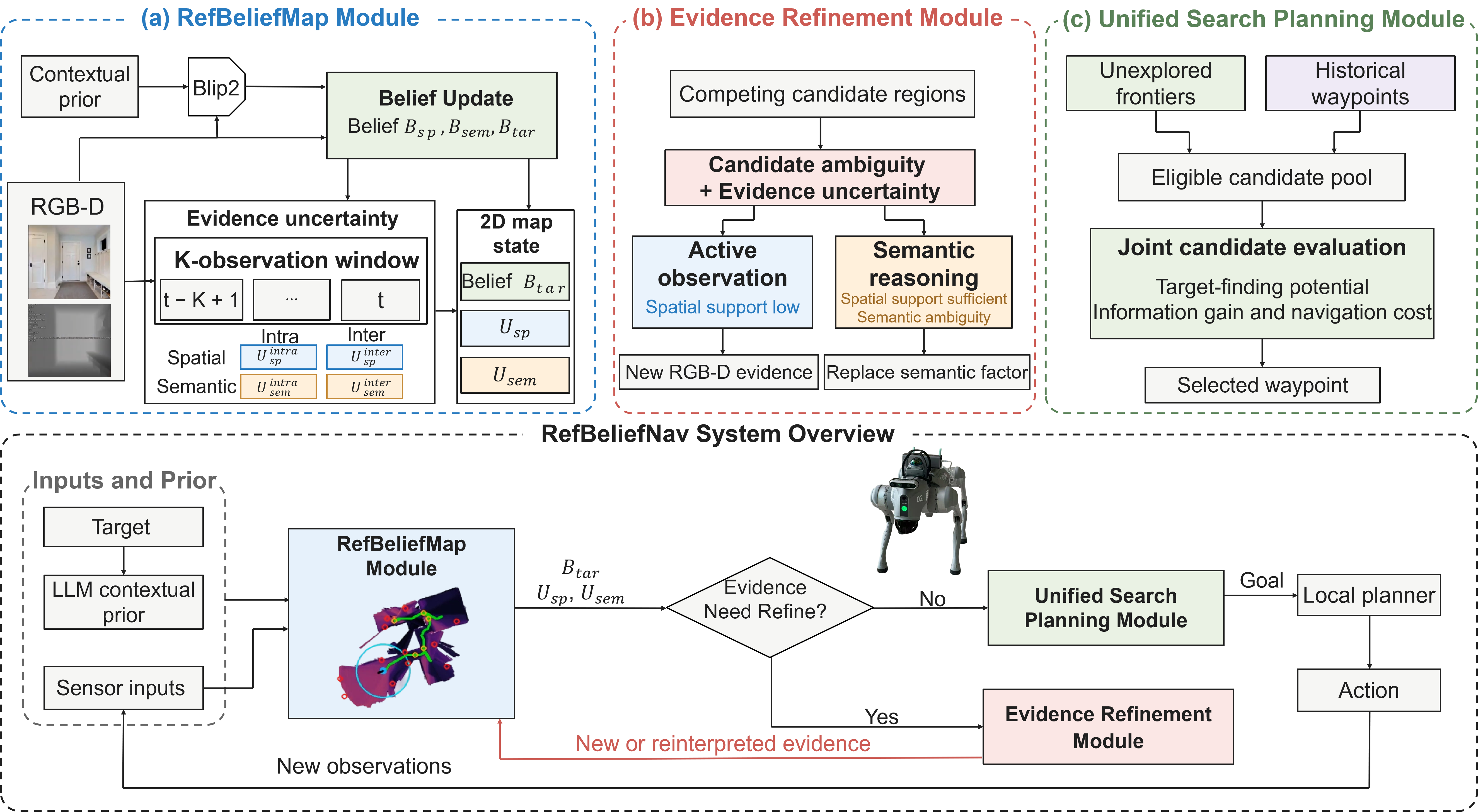}
    \caption{
        Overview of RefBeliefNav.
RefBeliefNav comprises three components: RefBeliefMap models
target relevance and evidence uncertainty; evidence refinement
improves insufficient support through active observation and
semantic reasoning; unified search planning selects among
frontiers and historical waypoints.
    }
    \label{fig:F1}
\end{figure*}

\label{sec:system_overview}

RefBeliefNav is an uncertainty-guided zero-shot object navigation framework that refines the evidence supporting target-location beliefs before navigation commitment. As illustrated in Fig.~\ref{fig:F1}, the framework consists of three modules:
(1) \textbf{RefBeliefMap}, which represents both where the target is likely to be and how reliably the available spatial and semantic evidence supports these beliefs, providing a shared basis for refinement and planning, details see in Sec.~\ref{sec:3.3};
(2) \textbf{Evidence Refinement}, which identifies when the current evidence is insufficient for a navigation decision and selectively resolves spatial or semantic uncertainty, helping the agent distinguish promising search regions before committing to a goal, details see in Sec.~\ref{sec:3.4};
(3) \textbf{Unified Search Planning}, which determines where to search next by jointly considering unexplored frontiers and previously visited locations, coordinating exploration and revisitation to balance target discovery, uncertainty reduction, and navigation cost, details see in Sec.~\ref{sec:3.5}.

\subsection{RefBeliefMap Module}
\label{sec:3.3}
RefBeliefMap is a 2D grid map in which each cell maintains three
belief quantities, $(B_{\mathrm{tar}}, B_{\mathrm{sem}},
B_{\mathrm{sp}})$, and two uncertainty measures,
$(U_{\mathrm{sp}}, U_{\mathrm{sem}})$.
The target-location belief $B_{\mathrm{tar}}$ is jointly determined
by the semantic belief $B_{\mathrm{sem}}$ and the spatial belief
$B_{\mathrm{sp}}$. Their definitions and update rules are presented
in Sec.\ref{sec:3.3.1}.
The uncertainty measures $U_{\mathrm{sp}}$ and $U_{\mathrm{sem}}$
characterize the reliability of the evidence supporting the
spatial and semantic beliefs, respectively. Both account for
current and historical observations, as detailed in
Sec.\ref{sec:3.3.2}.

\subsubsection{Belief Representation}
\label{sec:3.3.1}

For a target category $c$, RefBeliefMap represents observational
evidence through a spatial belief $B_{\mathrm{sp}}$ and a semantic
belief $B_{\mathrm{sem}}$, which jointly update the target-location
belief $B_{\mathrm{tar}}$.
Below, we describe a single spatially grounded RGB-D observation
region; its spatial support and semantic scores refer to the
same region.

\paragraph{Spatial Belief.}
We model spatial support on the 2D map using a
visibility-gated, camera-aligned Gaussian:
\begin{equation}
B_{\mathrm{sp},t}(\mathbf{q})
=
\chi_t(\mathbf{q})
\exp\!\left[
-\frac{(r_t(\mathbf{q})-\mu_{r,t})^2}
      {2\sigma_{r,t}^{2}}
-\frac{(a_t(\mathbf{q})-\mu_{a,t})^2}
      {2\sigma_{a,t}^{2}}
\right].
\label{eq:spatial_belief}
\end{equation}
Here, $r_t(\mathbf{q})$ and $a_t(\mathbf{q})$ are the
signed lateral and longitudinal coordinates relative to
the camera. The means are determined by the support center
obtained from depth and camera pose, while
$\sigma_{r,t},\sigma_{a,t}>0$ control the support extents.
The ray-based visibility gate
$\chi_t(\mathbf{q})\in\{0,1\}$ excludes invalid or occluded
observation associations.
Thus, $B_{\mathrm{sp},t}(\mathbf{q})\in[0,1]$
measures spatial support rather than target-presence
probability.
Coordinate definitions, visibility gating, and peak
normalization are detailed in
Appendix~\ref{app:spatial_belief}.

\paragraph{Semantic Belief.}
Given $c$, an LLM generates a set of target-conditioned prompts
$\mathcal{P}_c=\{p_m(c)\}_{m=1}^{M}$.
These prompts retain the same target category while varying
its descriptions and associated-object cues.
Let $f_I$ and $f_T$ denote the Blip-2 \cite{Blip-2} image and text encoders.
For the observed RGB region $I_t^{\mathrm{rgb}}$, the semantic
relevance under prompt $p_m(c)$ is
\begin{equation}
s_{t,m}
=
\frac{
f_I(I_t^{\mathrm{rgb}})^{\top}f_T(p_m(c))
}{
\|f_I(I_t^{\mathrm{rgb}})\|_2
\|f_T(p_m(c))\|_2
}.
\label{eq:semantic_similarity}
\end{equation}

We approximate the distribution of prompt-conditioned scores
with a Gaussian whose parameters are
$
\mu_{\mathrm{sem},t}
= \frac{1}{M}\sum_{m=1}^{M}s_{t,m},
\sigma_{\mathrm{sem},t}^{2}
= \frac{1}{M}\sum_{m=1}^{M}
(s_{t,m}-\mu_{\mathrm{sem},t})^2+\varepsilon^2,
$
where $\varepsilon>0$ provides a positive variance floor.
For cells associated with this observation region, the
single-step semantic belief is
$
B_{\mathrm{sem},t}(\mathbf{q})
=
\mathcal{N}
\bigl(\,\cdot\,;
\mu_{\mathrm{sem},t},\sigma_{\mathrm{sem},t}^{2}\bigr).
$ The mean summarizes target relevance, whereas the variance
captures sensitivity to the prompt set.

\paragraph{Target-Belief Update.}
Let $b(\mathbf{q})$ denote the latent target-relevance score
at location $\mathbf{q}$.
We represent its accumulated belief as
$
B_{\mathrm{tar},t}(\mathbf{q})
=
\mathcal{N}
\bigl(\,\cdot\,;
\mu_{\mathrm{tar},t}(\mathbf{q}),
\sigma_{\mathrm{tar},t}^{2}(\mathbf{q})\bigr),
$
with an initial prior specified by
$\mu_{\mathrm{tar},0}(\mathbf{q})$ and
$\sigma_{\mathrm{tar},0}^{2}(\mathbf{q})>0$.
To couple spatial support with semantic evidence, we adopt
a spatially weighted Gaussian observation model:
$
\mu_{\mathrm{sem},t}\mid b(\mathbf{q})
\sim \mathcal{N}\bigl(b(\mathbf{q}),R_t(\mathbf{q})\bigr), 
R_t(\mathbf{q})
= \frac{\sigma_{\mathrm{sem},t}^{2}}
         {B_{\mathrm{sp},t}(\mathbf{q})},
$

for $B_{\mathrm{sp},t}(\mathbf{q})>0$.
Stronger spatial support increases the observation precision. Under a conditionally independent observation approximation,
Gaussian fusion gives
\begin{equation}
\begin{aligned}
\sigma_{\mathrm{tar},t}^{2}(\mathbf{q})
&=
\left[
\frac{1}{\sigma_{\mathrm{tar},t-1}^{2}(\mathbf{q})}
+
\frac{B_{\mathrm{sp},t}(\mathbf{q})}
     {\sigma_{\mathrm{sem},t}^{2}}
\right]^{-1},\\
\mu_{\mathrm{tar},t}(\mathbf{q})
&=
\sigma_{\mathrm{tar},t}^{2}(\mathbf{q})
\left[
\frac{\mu_{\mathrm{tar},t-1}(\mathbf{q})}
     {\sigma_{\mathrm{tar},t-1}^{2}(\mathbf{q})}
+
\frac{B_{\mathrm{sp},t}(\mathbf{q})\mu_{\mathrm{sem},t}}
     {\sigma_{\mathrm{sem},t}^{2}}
\right].
\end{aligned}
\label{eq:target_belief_update}
\end{equation}
When $B_{\mathrm{sp},t}(\mathbf{q})=0$, no observation is
incorporated and the previous target belief is retained.
The spatial and semantic evidence beliefs are computed from
each incoming observation, while $B_{\mathrm{tar}}$
accumulates their contributions over time.

\subsubsection{Evidence Uncertainty}
\label{sec:3.3.2}

We assess spatial uncertainty as insufficient visible
support and semantic uncertainty as disagreement intra
and inter observations.
These quantities characterize evidence reliability.

\paragraph{Intra Observation Evidence Uncertainty.}
Since $B_{\mathrm{sp},t}$ incorporates both Gaussian
support and ray-based visibility, we define
$
U_{\mathrm{sp},t}^{\mathrm{intra}}(\mathbf{q})
=1-B_{\mathrm{sp},t}(\mathbf{q}).
\label{eq:spatial_intra_uncertainty}
$
Semantic uncertainty measures the dispersion of
similarity scores across the $M$ target-conditioned
prompts:
$
U_{\mathrm{sem},t}^{\mathrm{intra}}(\mathbf{q})
=\sigma_{\mathrm{sem},t}^{2}
=\frac{1}{M}\sum_{m=1}^{M}
(s_{t,m}-\mu_{\mathrm{sem},t})^2+\varepsilon^2.
$
These statistics are associated with $\mathbf{q}$
only when $\chi_t(\mathbf{q})=1$.

\paragraph{Inter Observation Evidence Uncertainty.}
We maintain the latest $K$ observations and identify
those providing valid evidence for each location:
$
\mathcal{W}_t
=\{o_i\mid\max(1,t-K+1)\leq i\leq t\}, \:
\mathcal{I}_t(\mathbf{q})
=\{i\mid o_i\in\mathcal{W}_t,\,
\chi_i(\mathbf{q})=1\},\:
n_t(\mathbf{q}) =|\mathcal{I}_t(\mathbf{q})|.
$ 

\textbf{(1) Spatial Evidence Uncertainty.} Historical uncertainty measures
remaining support deficit rather than disagreement:
\begin{equation}
U_{\mathrm{sp},t}^{\mathrm{inter}}(\mathbf{q})
=1-\max_{\max(1,t-K+1)\leq i<t}
B_{\mathrm{sp},i}(\mathbf{q}),
\label{eq:spatial_inter_uncertainty}
\end{equation}
where an empty maximum is defined as zero.
Combining current and historical support gives
\begin{equation}
\begin{aligned}
U_{\mathrm{sp},t}(\mathbf{q})
&=\min\!\left\{
U_{\mathrm{sp},t}^{\mathrm{intra}}(\mathbf{q}),
U_{\mathrm{sp},t}^{\mathrm{inter}}(\mathbf{q})
\right\}
=1-\max_{\max(1,t-K+1)\leq i\leq t}
B_{\mathrm{sp},i}(\mathbf{q}).
\end{aligned}
\label{eq:total_spatial_uncertainty}
\end{equation}

\textbf{(2) Semantic Evidence Uncertainty.} Only observations associated
with the same local support and target category
are compared.
For $n_t(\mathbf{q})>0$, define
\begin{equation}
\begin{aligned}
\bar{\mu}_{\mathrm{sem},t}(\mathbf{q})
&=\frac{1}{n_t(\mathbf{q})}
\sum_{i\in\mathcal{I}_t(\mathbf{q})}
\mu_{\mathrm{sem},i},\quad
U_{\mathrm{sem},t}^{\mathrm{inter}}(\mathbf{q})
=\frac{1}{n_t(\mathbf{q})}
\sum_{i\in\mathcal{I}_t(\mathbf{q})}
\left(\mu_{\mathrm{sem},i}
-\bar{\mu}_{\mathrm{sem},t}(\mathbf{q})\right)^2.
\end{aligned}
\label{eq:inter_observation_uncertainty}
\end{equation}
The total semantic uncertainty is
$
U_{\mathrm{sem},t}(\mathbf{q})
=\frac{1}{n_t(\mathbf{q})}
\sum_{i\in\mathcal{I}_t(\mathbf{q})}
U_{\mathrm{sem},i}^{\mathrm{intra}}(\mathbf{q})
+U_{\mathrm{sem},t}^{\mathrm{inter}}(\mathbf{q}).
$

\subsection{Evidence Refinement Module}
\label{sec:3.4}

\subsubsection{Trigger mechanism}
\label{sec:3.4.1}

We trigger evidence refinement when a promising local
candidate lacks sufficient support or competes with
another uncertain candidate.
Let $\mathcal{C}_t$ contain reachable waypoints within
radius $R$ of the agent with valid evidence in
RefBeliefMap.
Spatially redundant candidates are merged before comparison.
For each waypoint $w$, let $\mathbf{q}_w$ denote its
associated evidence location. We define
\begin{equation}
\begin{aligned}
S_t(w)
&=\mu_{\mathrm{tar},t}(\mathbf{q}_w),
H_t(w)
=\mathbb{I}\!\left[
U_{\mathrm{sp},t}(\mathbf{q}_w)>\tau_{\mathrm{sp}}
\ \lor\
U_{\mathrm{sem},t}(\mathbf{q}_w)>\tau_{\mathrm{sem}}
\right],
\end{aligned}
\label{eq:refinement_candidate}
\end{equation}
where $S_t$ measures target relevance and $H_t$
indicates insufficient evidence.
Unavailable semantic uncertainty is treated as
insufficient evidence rather than zero uncertainty.
Let $w_{(1)}$ and $w_{(2)}$ be the highest-scoring
candidates, with
$\Delta_t=S_t(w_{(1)})-S_t(w_{(2)})$.
Refinement is triggered when
\begin{equation}
\begin{aligned}
T_t
={}&\mathbb{I}\!\left[S_t(w_{(1)})\geq\tau_{\mathrm{rel}}\right] 
\cdot\mathbb{I}\!\left[
H_t(w_{(1)})=1
\ \lor\
\left(\Delta_t\leq\delta
\land H_t(w_{(2)})=1\right)
\right].
\end{aligned}
\label{eq:refinement_trigger}
\end{equation}

\subsubsection{Active Observation}
\label{sec:3.4.2}

When spatial support is insufficient, we select a
viewpoint that jointly observes the local candidates
under a bounded motion cost.
Let
$\mathcal{Q}_t=\{\mathbf{q}_w\mid w\in\mathcal{C}_t\}$
denote their associated evidence locations.
This set is fixed during each viewpoint-selection step,
so candidate views are evaluated against the same
observation targets. A candidate view is specified by
$v=(\mathbf{p}_v,\theta_v)$, comprising a planar camera
position and heading.
Using the current occupancy map, we predict visibility as
\begin{equation}
\begin{aligned}
\widehat{\chi}_t(\mathbf{q};v)
={}&
\mathbb{I}\!\left[
d_{\min}\leq\|\mathbf{q}-\mathbf{p}_v\|_2\leq d_{\max}
\right]
\cdot
\mathbb{I}\!\left[
|\alpha_v(\mathbf{q})|\leq\phi_{\mathrm{obs}}/2
\right]
\prod_{\mathbf{u}\in\mathcal{R}_v(\mathbf{q})}
\mathbb{I}\!\left[\mathbf{u}\in\mathcal{F}_t\right],
\end{aligned}
\label{eq:predicted_view_visibility}
\end{equation}
where $\alpha_v(\mathbf{q})$ is the wrapped bearing
relative to $\theta_v$, $[d_{\min},d_{\max}]$ is the
observation distance range, and $\phi_{\mathrm{obs}}$
does not exceed the camera's horizontal field of view.
The ray cells $\mathcal{R}_v(\mathbf{q})$ exclude the
camera and endpoint cells, and $\mathcal{F}_t$ contains
known-free cells. We assign each evidence location a positive weight
$
\omega_t(\mathbf{q})
=
\frac{1+U_{\mathrm{sp},t}(\mathbf{q})}
{\sum_{\mathbf{q}'\in\mathcal{Q}_t}
[1+U_{\mathrm{sp},t}(\mathbf{q}')]}.
\label{eq:observation_weights}
$
The constant term encourages joint coverage, while
spatial uncertainty prioritizes insufficiently
supported locations.
The selected view solves
\begin{equation}
\begin{aligned}
v_t^\star\in
\operatorname*{arg\,max}_{v\in\mathcal{P}_t}
\quad&
\sum_{\mathbf{q}\in\mathcal{Q}_t}
\omega_t(\mathbf{q})
\widehat{\chi}_t(\mathbf{q};v)
\text{s.t.}\quad
c_t(v)\leq C_{\max},
\end{aligned}
\label{eq:active_view_optimization}
\end{equation}
where $\mathcal{P}_t$ is a finite set of collision-free,
reachable camera poses, and $c_t(v)$ accounts for
translation and rotation from the current pose.
Ties are resolved by the smaller motion cost.
The selected viewpoints are subsequently converted into
simulator-executable actions by an action converter,
as detailed in the Appendix \ref{app:action transitions}.

\subsubsection{Semantic Reasoning}
\label{sec:3.4.3}

Semantic reasoning reassesses visible regions with sufficient
spatial support but high semantic uncertainty within the
triggered neighborhood, while keeping the robot pose fixed.
A multimodal LLM evaluates current crops and their BLIP-2
scores (intra-observation) and checks consistency with
comparable historical observations of the same region
(inter-observation).
The prompts are provided in Appendix~\ref{app:prompt}.

For each region $j$, the two stages produce reliability
scores $r_{t,j}^{\mathrm{intra}},
r_{t,j}^{\mathrm{inter}}\in\{1,\ldots,10\}$,
with higher scores indicating stronger support.
We set
$\gamma_{t,j}
=\min(r_{t,j}^{\mathrm{intra}},
r_{t,j}^{\mathrm{inter}})/10$,
or $\gamma_{t,j}=r_{t,j}^{\mathrm{intra}}/10$
when comparable history is unavailable.
This heuristic reliability weight tempers the semantic
belief for each associated cell $\mathbf{q}$:
\begin{equation}
\widetilde{B}_{\mathrm{sem},t}(\mathbf{q})
=
\mathcal{N}\!\bigl(
\cdot;\mu_{\mathrm{sem},t},
\tfrac{\sigma_{\mathrm{sem},t}^{2}}{\gamma_{t,j}}
\bigr).
\label{eq:refined_semantic_belief}
\end{equation}
This downweights unreliable evidence without changing its
mean. The adjusted variance is used for target-belief fusion
(Sec.~\ref{sec:3.3.1}).

\subsection{Unified Search Planning Module}
\label{sec:3.5}

Using the refined RefBeliefMap, we jointly evaluate unexplored
frontiers and historical waypoints selected for their potential
revisitation value. Frontier generation follows
VLFM~\cite{yokoyama2024vlfm}, while the filtering criteria for
historical waypoints are detailed in
Appendix~\ref{app:rules for waypoint}.

 $\mathcal{V}_t^{\mathrm{fr}}$ and
$\mathcal{V}_t^{\mathrm{hist}}$ denote the corresponding
reachable candidate sets, and let
$\mathcal{V}_t=
\mathcal{V}_t^{\mathrm{fr}}\cup
\mathcal{V}_t^{\mathrm{hist}}$.
The next waypoint is selected by
\begin{equation}
w_t^\star
\in
\operatorname*{arg\,max}_{w\in\mathcal{V}_t}
\left[
J_{\mathrm{tar},t}(w)
+\lambda_{\mathrm{unc}}J_{\mathrm{unc},t}(w)
-\lambda_{\mathrm{cost}}c_t(w)
\right],
\label{eq:unified_search_objective}
\end{equation}
where the three terms represent target relevance score,
potential information benefit, and navigation cost,
respectively, and the weights are nonnegative.
\paragraph{Ray-based Target-Relevance Score.}
For a candidate $w$ at $\mathbf{p}_w$, we model a local
$2\pi$ yaw scan using uniformly spaced 2D rays within
sensing range, terminating at and including the first
occupied or unknown cell.
Let $\mathcal{E}_t$ denote cells with accumulated target
evidence, and let $\mathcal{O}_t(w)$ contain the unique
ray-reached cells in $\mathcal{E}_t$ together with first
unknown endpoints.
Using the distance weight
$\kappa_w(\mathbf{q})
=\exp(-\|\mathbf{q}-\mathbf{p}_w\|_2/\ell)$
with $\ell>0$, we compute
$
J_{\mathrm{tar},t}(w)
=
\sum_{\mathbf{q}\in\mathcal{O}_t(w)\cap\mathcal{E}_t}
\kappa_w(\mathbf{q})\mu_{\mathrm{tar},t}(\mathbf{q}).
$

\paragraph{Uncertainty-driven Information Benefit.}
We use the same ray footprint to identify locations
where additional observations may improve the evidence.
Since $U_{\mathrm{sp},t}\in[0,1]$ while
$U_{\mathrm{sem},t}$ is a score variance, we first
normalize semantic uncertainty:
$\bar{U}_{\mathrm{sem},t}(\mathbf{q})
=U_{\mathrm{sem},t}(\mathbf{q})/
[U_{\mathrm{sem},t}(\mathbf{q})+\tau_{\mathrm{sem}}]$

using the positive semantic threshold as a scale.
The information-benefit score is
\begin{equation}
\begin{aligned}
J_{\mathrm{unc},t}(w)
=
\sum_{\mathbf{q}\in\mathcal{O}_t(w)}
\kappa_w(\mathbf{q})
\Big[
&U_{\mathrm{sp},t}(\mathbf{q})
+
\big(1-U_{\mathrm{sp},t}(\mathbf{q})\big)
\bar{U}_{\mathrm{sem},t}(\mathbf{q})
\Big].
\end{aligned}
\label{eq:ray_uncertainty_score}
\end{equation}
The first term rewards opportunities to improve
insufficient spatial support.
The second emphasizes semantic ambiguity where
spatial evidence is already supported.

\paragraph{Navigation Cost.}
We use A$^\ast$\cite{A} to compute a collision-free path from
 agent position $\mathbf{p}_t$ to the candidate
position $\mathbf{p}_w$ on the 
map, and define its length as the navigation cost
$c_t(w)$.

\section{Experimental Results}
\subsection{Benchmarks and Implementation details}
\noindent\textbf{Datasets.}
We evaluate our method on HM3D~\cite{HM3D} and
MP3D~\cite{MP3D} using the standard Object Goal Navigation
(OGN) benchmark. The HM3D validation split comprises
2,000 episodes across 20 scenes and six object categories,
while the MP3D validation split includes 2,195 episodes
across 11 scenes and 21 categories.

\noindent\textbf{Evaluation Metrics.}
We report two standard metrics: Success Rate (SR) and
Success weighted by inverse Path Length (SPL).
SR measures the proportion of episodes completed
successfully, whereas SPL evaluates navigation efficiency
by weighting successful episodes according to the ratio
between the shortest-path distance and the executed
trajectory length. 

\noindent\textbf{Implementation Details.}
We set
$K=5$, $M=8$,
$\sigma_{r,t}=0.50\,\mathrm{m}$,
$\sigma_{a,t}=1.00\,\mathrm{m}$,
and $\varepsilon=0.02$.
The initial target belief uses
$\mu_{\mathrm{tar},0}(\mathbf{q})=0$
and $\sigma_{\mathrm{tar},0}^{2}(\mathbf{q})=1$.
For refinement triggering, we use
$R=3.0\,\mathrm{m}$,
$\delta=0.03$,
$\tau_{\mathrm{rel}}=0.25$,
$\tau_{\mathrm{sp}}=0.50$,
and $\tau_{\mathrm{sem}}=0.01$.
For active observation, we choose
$d_{\min}=0.50\,\mathrm{m}$,
$d_{\max}=3.0\,\mathrm{m}$,
$\phi_{\mathrm{obs}}=\phi_{\mathrm{cam}}$,
and $C_{\max}=8$,
where $\phi_{\mathrm{cam}}$ denotes the fixed horizontal
camera field of view.
The viewpoint cost
$c_t(v)$ counts discrete translation and rotation actions,
so $C_{\max}$ is expressed in actions.
For unified planning, we set
$\ell=3.0\,\mathrm{m}$,
$\lambda_{\mathrm{unc}}=0.25$,
and $\lambda_{\mathrm{cost}}=5.0\,\mathrm{m}^{-1}$,
while $c_t(w)$ remains the A* path length in meters.
The reliability factor $\gamma_{t,j}$ remains
observation-dependent and is not assigned a fixed default. On the selection of General Models, We use Grounding-DINO~\cite{Gd} for open-vocabulary object
detection and MobileSAM~\cite{mobile_sam} for object segmentation.
we employ BLIP-2~\cite{Blip-2} for semantic similarity
representations and GPT-4o for high-level reasoning.
All experiments are conducted online on a workstation
equipped with a single NVIDIA RTX 4090 GPU. 

\subsection{Method Performance}

\begin{table*}[t]
\centering
\caption{Comparison of RefBeliefNav and zero-shot methods on the HM3D and MP3D datasets.}
\label{tab:comparison}
\setlength{\tabcolsep}{7pt}
\begin{tabular}{l c cc cc}
\toprule
\multirow{2}{*}{Method}
& \multirow{2}{*}{Venue}
& \multicolumn{2}{c}{HM3D}
& \multicolumn{2}{c}{MP3D} \\
\cmidrule(lr){3-4} \cmidrule(lr){5-6}
& & SR$\uparrow$ & SPL$\uparrow$ & SR$\uparrow$ & SPL$\uparrow$ \\
\midrule
ZSON \cite{Zson}
& NeurIPS'22 & 25.5 & 12.6 & 15.3 & 4.8 \\
ESC \cite{zhou2023esc}
& ICML'23 & 39.2 & 22.3 & 28.7 & 14.2 \\
L3MVN \cite{L3mvn}
& IROS'23 & 50.4 & 23.1 & 34.9 & 14.5 \\
PSL \cite{PSL}
& ECCV'24 & 42.4 & 19.2 & 18.9 & 6.9 \\
VLFM \cite{yokoyama2024vlfm}
& ICRA'24 & 52.5 & 30.4 & 36.4 & 17.5 \\
OpenFMNav \cite{kuang2024openfmnav}
& NAACL'24 & 52.5 & 24.1 & 37.2 & 15.7 \\
SG-NAV \cite{yin2024sgnav}
& NeurIPS'24 & 54.0 & 24.9 & 40.2 & 16.0 \\
BeliefMapNav \cite{zhou2025beliefmapnav}
& NeurIPS'25 & 61.4 & 30.6 & 37.3 & 17.6 \\
ASCENT \cite{Stairway}
& RAL'26 & 65.4 & 33.5 & 44.5 & 15.5 \\
\midrule
\textbf{RefBeliefNav}
& \textbf{ours}
& \textbf{69.2}
& \textbf{35.3}
& \textbf{49.1}
& \textbf{18.5} \\
\bottomrule
\end{tabular}
\end{table*}

Table~\ref{tab:comparison} compares our RefBeliefNav with representative zero-shot object navigation methods on HM3D and MP3D, descriptions of these baselines are provided in the Appendix \ref{app:baseline description}. Our RefBeliefNav achieves the highest SR and SPL among the compared methods on both benchmarks. On HM3D, it obtains an SR of 69.2\% and an SPL of 35.3\%, exceeding ASCENT\cite{Stairway}, the strongest baseline on both metrics, by 3.8 and 1.8 percentage points. On MP3D, RefBeliefNav achieves an SR of 49.1\%, surpassing ASCENT\cite{Stairway} by 4.6 percentage points. Overall, these results demonstrate consistent gains in SR and SPL efficiency across both datasets.

\subsection{Ablative Study}
\label{sec:4.3}
In order to efficiently evaluate the contribution of each module in our method and the sensitivity of the parameters, we randomly sampled and tested 1000 navigation segments from HM3D, using a fixed random seed.
We present ablation results for the key modules in the main text,
with additional ablation studies and parameter sensitivity analyses
provided in the Appendix \ref{sec:Parameter Sensitivity Analyses}.

\begin{table}[htbp]
    \centering

    \begin{minipage}[t]{0.48\linewidth}
        \vspace{0pt}
        \centering
        \caption{Impact of the Evidence Refinement Module.}
        \label{tab:evidence_refinement_ablation}
        \setlength{\tabcolsep}{5pt}

        \resizebox{\linewidth}{!}{%
            \begin{tabular}{lcc}
                \toprule
                Method & SR$\uparrow$ & SPL$\uparrow$ \\
                \midrule
                 w/o semantic evidence refinement
                & 65.2 & 33.9 \\
                 w/o spatial evidence refinement
                & 64.5 & 33.7 \\
                 w/o evidence refinement
                & 61.4 & 31.3 \\
                \midrule
                \textbf{RefBeliefNav}
                & \textbf{69.5} & \textbf{35.5} \\
                \bottomrule
            \end{tabular}%
        }
    \end{minipage}\hfill
    \begin{minipage}[t]{0.48\linewidth}
        \vspace{0pt}
        \centering
        \caption{Ablation of Unified Search Planning.}
        \label{tab:planning_ablation}
        \setlength{\tabcolsep}{5pt}

        \resizebox{\linewidth}{!}{%
            \begin{tabular}{lcc}
                \toprule
                Method & SR$\uparrow$ & SPL$\uparrow$ \\
                \midrule
                Frontier-only candidates
                & 64.6 & 33.9 \\
                w/o information benefit term
                & 58.2 & 33.8 \\
                w/o navigation cost
                & 66.7 & 30.4 \\
                w/o target-relevance score
                & 56.1 & 31.2 \\
                \midrule
                \textbf{RefBeliefNav}
                & \textbf{69.5} & \textbf{35.5} \\
                \bottomrule
            \end{tabular}%
        }
    \end{minipage}
\end{table}

\paragraph{Effectiveness of Evidence Refinement Module.}
Table~\ref{tab:evidence_refinement_ablation} shows that
removing either semantic or spatial refinement lowers
SR and SPL. Disabling both yields the largest drops
of 8.1 and 4.2 percentage points, respectively,
supporting the joint use of both mechanisms.

\paragraph{Effectiveness of Unified Search Planning.}
Table~\ref{tab:planning_ablation} shows that removing
target relevance or information benefit reduces SR
to 56.1\% or 58.2\%, respectively, highlighting their
complementary roles in balancing exploration and exploitation.
Removing navigation cost lowers SPL from 35.5\% to 30.4\%,
while frontier-only candidates reduce SR to 64.6\%,
supporting cost-aware planning and the use of historical
waypoints. The full planner achieves the highest SR and SPL.

\subsection{Qualitative Analysis}
\label{sec:qualitative_analysis}

\begin{figure}[t]
    \centering
    \includegraphics[width=\linewidth]{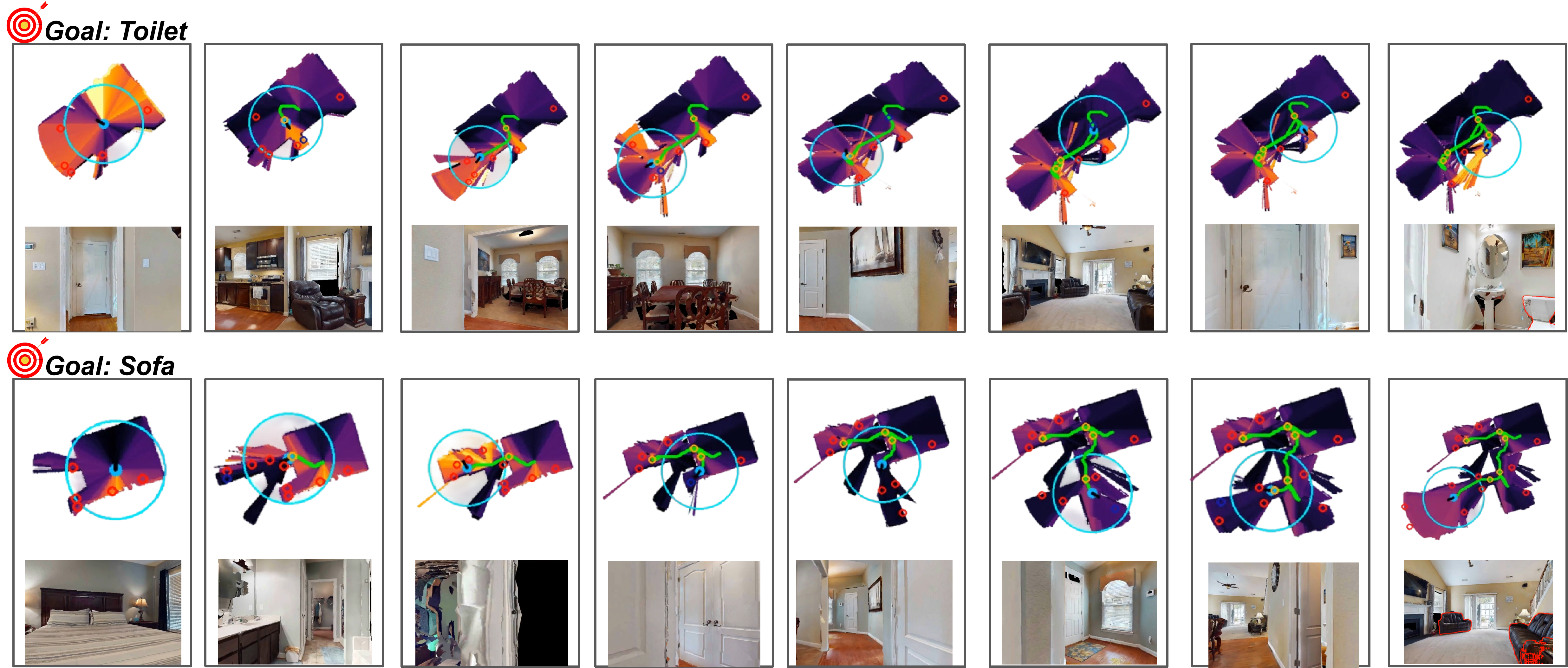}
    \caption{
    Qualitative examples of RefBeliefNav searching for a toilet
    (top) and a sofa (bottom) in simulation.
    Each panel shows the evolving RefBeliefMap and the corresponding
    egocentric RGB observation.
    Orange, red, and dark blue hollow circles denote historical
    waypoints, frontiers, and the selected waypoint, respectively.
    The light blue circle marks the radius-$R$ neighborhood centered
    on the robot used to evaluate refinement triggers
    (Sec. \ref{sec:3.4.1}).
    }
    \label{fig:qualitative_sim}
\end{figure}

\begin{figure}[t]
    \centering
    \includegraphics[width=\linewidth]{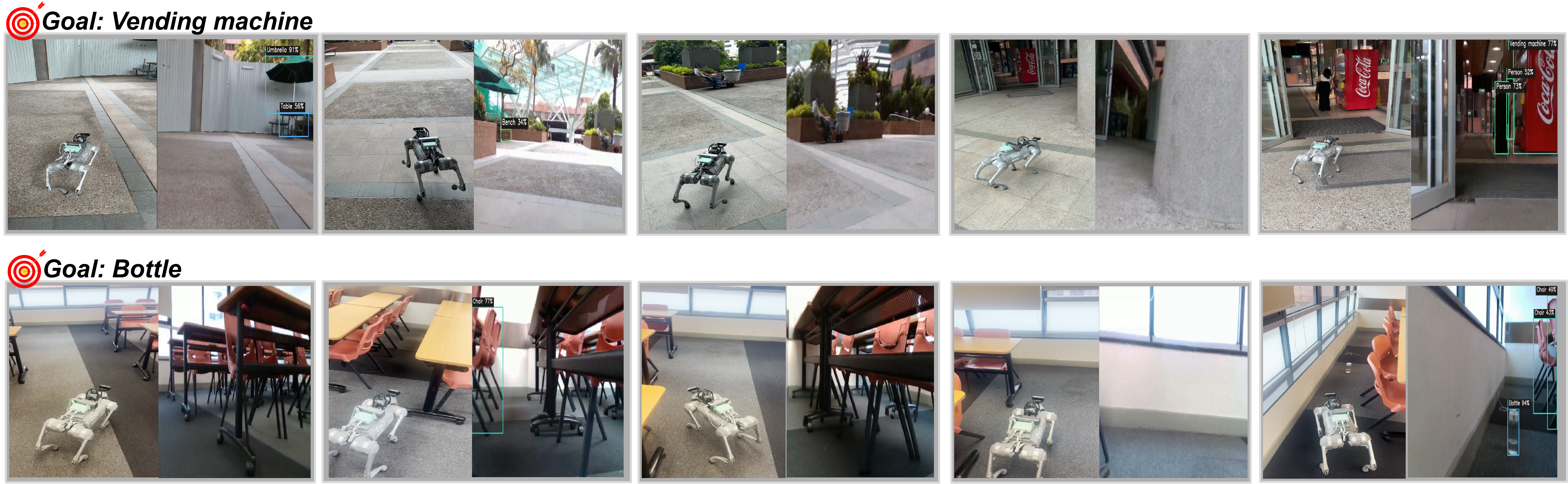}
    \caption{
Real-world searches for a vending machine in an open public
area (top) and a bottle in a classroom (bottom).
The sequences progress from left to right, with each panel
pairing an external view of the robot with its egocentric
RGB observation. Bounding boxes indicate object detections.
    }
    \label{fig:qualitative_real}
\end{figure}

In simulation (Fig.~\ref{fig:qualitative_sim}), RefBeliefNav
jointly considers target beliefs and spatial and semantic
evidence uncertainty to guide exploration and revisitation.
Real-world deployment on the Unitree Go2
(Fig.~\ref{fig:qualitative_real}) further illustrates how
position and viewpoint adjustments reveal initially unseen
targets under clutter and limited visibility.
Additional examples and complete navigation sequences are
provided in the supplementary video.

\section{Conclusion}
We presented RefBeliefNav, an uncertainty-guided framework
for zero-shot object navigation. Its RefBeliefMap distinguishes
target beliefs from spatial and semantic evidence uncertainty,
guiding active observation and semantic reasoning to refine
the evidence supporting navigation decisions. Joint evaluation
of frontiers and historical waypoints unifies exploration
and revisitation. Experiments on HM3D and MP3D demonstrate
higher SR and SPL than the evaluated baselines, while
deployment on a Unitree Go2 demonstrates real-world
applicability.

\section*{AI use strategyment}
In this work, we used generative AI tools (e.g., large language models) for manuscript language editing, grammar refinement, and code debugging. We have not used generative AI tools for generating core research ideas, designing methodologies, or creating experimental data. We have thoroughly reviewed all AI-assisted text and code to ensure their accuracy, correctness, and originality. We take full responsibility for the final content of this work, including all text, claims, and artifacts.

\bibliography{iclr2027_conference}

@inproceedings{yokoyama2024vlfm,
  title={Vlfm: Vision-language frontier maps for zero-shot semantic navigation},
  author={Yokoyama, Naoki and Ha, Sehoon and Batra, Dhruv and Wang, Jiuguang and Bucher, Bernadette},
  booktitle={2024 IEEE International Conference on Robotics and Automation (ICRA)},
  pages={42--48},
  year={2024},
  Xorganization={IEEE}
}

@inproceedings{kuang2024openfmnav,
  author = {Yuxuan Kuang and Hai Lin and Meng Jiang},
  title = {{OpenFMNav}: Towards Open-Set Zero-Shot Object Navigation via Vision-Language Foundation Models},
  booktitle = {Findings of the Association for Computational Linguistics: NAACL 2024},
  pages={338--351},
  year = {2024},
}

@inproceedings{cai2024pixnav,
  author = {Wenzhe Cai and Siyuan Huang and Guangran Cheng and Yuxing Long and Peng Gao and Changyin Sun and Hao Dong},
  booktitle={2024 IEEE International Conference on Robotics and Automation (ICRA)}, 
  title={Bridging Zero-shot Object Navigation and Foundation Models through Pixel-Guided Navigation Skill}, 
  year={2024},
  volume={},
  number={},
  pages={5228-5234},
}

@inproceedings{yokoyama2024hm3dovon,
  author = {Naoki Yokoyama and Ram Ramrakhya and Abhishek Das and Dhruv Batra and Sehoon Ha},
  booktitle={2024 IEEE/RSJ International Conference on Intelligent Robots and Systems (IROS)}, 
  title={HM3D-OVON: A Dataset and Benchmark for Open-Vocabulary Object Goal Navigation}, 
  year={2024},
  volume={},
  number={},
  pages={5543-5550},
}

@inproceedings{yuan2024gamap,
  author = {Shuaihang Yuan and Hao Huang and Yu Hao and Congcong Wen and Anthony Tzes and Yi Fang},
  title = {{GAMap}: Zero-Shot Object Goal Navigation with Multi-Scale Geometric-Affordance Guidance},
  booktitle = {Advances in Neural Information Processing Systems (NeurIPS)},
  pages = {39386--39408},
  volume = {37},
  year = {2024},
}

@inproceedings{zhou2025beliefmapnav,
  author = {Zibo Zhou and Yue Hu and Lingkai Zhang and Zonglin Li and Siheng Chen},
  title = {{BeliefMapNav}: {3D} Voxel-Based Belief Map for Zero-Shot Object Navigation},
  booktitle = {Advances in Neural Information Processing Systems (NeurIPS)},
  pages = {83008--83036},
  volume = {38},
  year = {2025},
}

@inproceedings{gu2024conceptgraphs,
  author = {Qiao Gu and Alihusein Kuwajerwala and Sacha Morin and Krishna Murthy Jatavallabhula and Bipasha Sen and Aditya Agarwal and Corban Rivera and William Paul and Kirsty Ellis and Rama Chellappa and Chuang Gan and Celso Miguel de Melo and Joshua B. Tenenbaum and Antonio Torralba and Florian Shkurti and Liam Paull},
  booktitle={2024 IEEE International Conference on Robotics and Automation (ICRA)}, 
  title={ConceptGraphs: Open-Vocabulary 3D Scene Graphs for Perception and Planning}, 
  year={2024},
  volume={},
  number={},
  pages={5021-5028},
}

@inproceedings{werby2024hovsg,
  author = {Abdelrhman Werby and Chenguang Huang and Martin B{\"u}chner and Abhinav Valada and Wolfram Burgard},
  title = {Hierarchical Open-Vocabulary {3D} Scene Graphs for Language-Grounded Robot Navigation},
  booktitle = {Proceedings of Robotics: Science and Systems (RSS)},
  year = {2024},
}

@inproceedings{yin2024sgnav,
  author = {Hang Yin and Xiuwei Xu and Zhenyu Wu and Jie Zhou and Jiwen Lu},
  title = {{SG-Nav}: Online {3D} Scene Graph Prompting for {LLM}-based Zero-shot Object Navigation},
  journal={Advances in Neural Information Processing Systems (NeurIPS)},
  volume={37},
  pages={5285--5307},
  year={2024}
}

@inproceedings{yin2025unigoal,
  author={Yin, Hang and Xu, Xiuwei and Zhao, Linqing and Wang, Ziwei and Zhou, Jie and Lu, Jiwen},
  booktitle={2025 IEEE/CVF Conference on Computer Vision and Pattern Recognition (CVPR)}, 
  title={UniGoal: Towards Universal Zero-shot Goal-oriented Navigation}, 
  year={2025},
  volume={},
  number={},
  pages={19057-19066},
}

@inproceedings{zhang2026glmap,
  author    = {Zhang, Sixian and Wang, Yiyao and Song, Xinhang and Zhang, Keming and Xu, Zijian and Jiang, Shuqiang},
  title     = {Multi-Scale Gaussian-Language Map for Zero-shot Embodied Navigation and Reasoning},
  booktitle = {Proceedings of the IEEE/CVF Conference on Computer Vision and Pattern Recognition (CVPR)},
  year      = {2026},
  pages     = {37086-37097}
}

@inproceedings{wu2024voronav,
  author = {Pengying Wu and Yao Mu and Bingxian Wu and Yi Hou and Ji Ma and Shanghang Zhang and Chang Liu},
  title = {{VoroNav}: Voronoi-based Zero-shot Object Navigation with Large Language Model},
  booktitle = {Proceedings of the International Conference on Machine Learning (ICML)},
  year = {2024},
  pages = {53737-53775}
}

@inproceedings{ginting2024seek,
  author = {Muhammad Fadhil Ginting and Sung-Kyun Kim and David D. Fan and Matteo Palieri and Mykel J. Kochenderfer and Ali-akbar Agha-Mohammadi},
  title = {{SEEK}: Semantic Reasoning for Object Goal Navigation in Real World Inspection Tasks},
  booktitle = {Proceedings of Robotics: Science and Systems (RSS)},
  year = {2024},
}

@inproceedings{zhang2025fbn,
  author={Zhang, Sixian and Yu, Xinyao and Song, Xinhang and Wang, Yiyao and Jiang, Shuqiang},
  booktitle={2025 IEEE/CVF International Conference on Computer Vision (ICCV)}, 
  title={Function-Centric Bayesian Network for Zero-Shot Object Goal Navigation}, 
  year={2025},
  volume={},
  number={},
  pages={19535-19545},
}

@inproceedings{zhang2024trihelper,
  author={Zhang, Lingfeng and Zhang, Qiang and Wang, Hao and Xiao, Erjia and Jiang, Zixuan and Chen, Honglei and Xu, Renjing},
  booktitle={2024 IEEE/RSJ International Conference on Intelligent Robots and Systems (IROS)}, 
  title={TriHelper: Zero-Shot Object Navigation with Dynamic Assistance}, 
  year={2024},
  volume={},
  number={},
  pages={10035-10042},
}

@inproceedings{cao2025cognav,
  author={Cao, Yihan and Zhang, Jiazhao and Yu, Zhinan and Liu, Shuzhen and Qin, Zheng and Zou, Qin and Du, Bo and Xu, Kai},
  booktitle={2025 IEEE/CVF International Conference on Computer Vision (ICCV)}, 
  title={CogNav: Cognitive Process Modeling for Object Goal Navigation with LLMs}, 
  year={2025},
  volume={},
  number={},
  pages={9550-9560},
}

@article{zhang2025apexnav,
  author={Zhang, Mingjie and Du, Yuheng and Wu, Chengkai and Zhou, Jinni and Qi, Zhenchao and Ma, Jun and Zhou, Boyu},
  journal={IEEE Robotics and Automation Letters (RAL)}, 
  title={ApexNAV: An Adaptive Exploration Strategy for Zero-Shot Object Navigation With Target-Centric Semantic Fusion}, 
  year={2025},
  volume={10},
  number={11},
  pages={11530-11537},
}

@inproceedings{nie2025wmnav,
  author={Nie, Dujun and Guo, Xianda and Duan, Yiqun and Zhang, Ruijun and Chen, Long},
  booktitle={2025 IEEE/RSJ International Conference on Intelligent Robots and Systems (IROS)}, 
  title={WMNav: Integrating Vision-Language Models into World Models for Object Goal Navigation}, 
  year={2025},
  volume={},
  number={},
  pages={2392-2399},
}

@inproceedings{chen2026chainofsearch,
  author = {Hanrui Chen and Liqi Yan and Qifan Wang and Jianhui Zhang and Fangli Guan and Pan Li},
  title = {{Chain-of-Search}: Parameter-Efficient Reasoning for Zero-Shot Object Navigation},
  booktitle = {Proceedings of the AAAI Conference on Artificial Intelligence},
  year = {2026},
  pages = {1721–1729},
}

@inproceedings{wang2026trajrag,
  author    = {Wang, Yiyao and Zhang, Sixian and Zhang, Keming and Song, Xinhang and Du, Songjie and Jiang, Shuqiang},
  title     = {TrajRAG: Retrieving Geometric-Semantic Experience for Zero-Shot Object Navigation},
  booktitle = {Proceedings of the IEEE/CVF Conference on Computer Vision and Pattern Recognition (CVPR)},
  year      = {2026},
  pages     = {15166-15176}
}

@inproceedings{niu2026ssmgnav,
  author = {Haochen Niu and Lantao Zhang and Xingwu Ji and Rendong Ying and Peilin Liu and Fei Wen},
  title = {{SSMG-Nav}: Enhancing Lifelong Object Navigation with Semantic Skeleton Memory Graph},
  booktitle = {IEEE International Conference on Robotics and Automation (ICRA)},
  year = {2026},
}

@inproceedings{ramakrishnan2022poni-train1,
  author={Ramakrishnan, Santhosh Kumar and Chaplot, Devendra Singh and Al-Halah, Ziad and Malik, Jitendra and Grauman, Kristen},
  booktitle={2022 IEEE/CVF Conference on Computer Vision and Pattern Recognition (CVPR)}, 
  title={PONI: Potential Functions for ObjectGoal Navigation with Interaction-free Learning}, 
  year={2022},
  volume={},
  number={},
  pages={18868-18878},
}

@inproceedings{train2,
  author={Ramrakhya, Ram and Batra, Dhruv and Wijmans, Erik and Das, Abhishek},
  booktitle={2023 IEEE/CVF Conference on Computer Vision and Pattern Recognition (CVPR)}, 
  title={PIRLNav: Pretraining with Imitation and RL Finetuning for OBJECTNAV}, 
  year={2023},
  volume={},
  number={},
  pages={17896-17906},
}

@inproceedings{train3,
  author={Fukushima, Rui and Ota, Kei and Kanezaki, Asako and Sasaki, Yoko and Yoshiyasu, Yusuke},
  booktitle={2022 International Conference on Robotics and Automation (ICRA)}, 
  title={Object Memory Transformer for Object Goal Navigation}, 
  year={2022},
  volume={},
  number={},
  pages={11288-11294},
}

@inproceedings{train4,
  author={Maksymets, Oleksandr and Cartillier, Vincent and Gokaslan, Aaron and Wijmans, Erik and Galuba, Wojciech and Lee, Stefan and Batra, Dhruv},
  booktitle={2021 IEEE/CVF International Conference on Computer Vision (ICCV)}, 
  title={THDA: Treasure Hunt Data Augmentation for Semantic Navigation}, 
  year={2021},
  volume={},
  number={},
  pages={15354-15363},
}

@article{train5,
  author={Ye, Xin and Yang, Yezhou},
  journal={IEEE Robotics and Automation Letters (RAL)}, 
  title={Efficient Robotic Object Search Via HIEM: Hierarchical Policy Learning With Intrinsic-Extrinsic Modeling}, 
  year={2021},
  volume={6},
  number={3},
  pages={4425-4432},
}

@inproceedings{Blip-2,
  title={Blip-2: Bootstrapping language-image pre-training with frozen image encoders and large language models},
  author={Li, Junnan and Li, Dongxu and Savarese, Silvio and Hoi, Steven},
  booktitle = {Proceedings of the 40th International Conference on Machine Learning (ICML)},
  pages={19730--19742},
  year={2023},
}

@inproceedings{Clip,
  title={Learning transferable visual models from natural language supervision},
  author={Radford, Alec and Kim, Jong Wook and Hallacy, Chris and Ramesh, Aditya and Goh, Gabriel and Agarwal, Sandhini and Sastry, Girish and Askell, Amanda and Mishkin, Pamela and Clark, Jack and others},
  booktitle= {Proceedings of the 38th International Conference on Machine Learning (ICML)},
  pages={8748--8763},
  year={2021},
  Xorganization={PMLR}
}

@inproceedings{ko2025care,
  title     = {Context-Aware Replanning with Pre-Explored Semantic Map for Object Navigation},
  author    = {Ko, Po-Chen and Su, Hung-Ting and Chen, CY and Yeh, Jia-Fong and Sun, Min and Hsu, Winston H.},
  booktitle = {Proceedings of the 8th Conference on Robot Learning},
  Xseries    = {Proceedings of Machine Learning Research},
  volume    = {270},
  pages     = {4253--4267},
  year      = {2025},
  Xpublisher = {PMLR},
  Xurl       = {https://proceedings.mlr.press/v270/ko25b.html}
}

@article{zhang2026igvrrt,
  title={IGV-RRT: Prior-Real-Time Observation Fusion for Active Object Search in Changing Environments}, 
  author={Wei Zhang and Ping Gong and Yujie Wang and Leilei Yao and Minghui Bai and Rongfeng Ye and Yinchuan Wang and Yachao Wang and Chen Sun and Chaoqun Wang},
  journal={arXiv preprint \href{https://arxiv.org/abs/2603.21887}{arXiv:2603.21887}},
  year={2026},
}

@article{liu2026aecnav,
  title         = {{AECNav}: Active Evidence Consolidation for Efficient Zero-Shot Open-Vocabulary Object Navigation},
  author        = {Liu, Guanlin and Ling, Shaobin and Liu, Renyuan and Gong, Zeying and Hu, Junjie},
  journal={arXiv preprint \href{https://arxiv.org/abs/2608.10817}{arXiv:2608.10817}},
  year          = {2026},
}

@article{xiao2026aerobelief,
  title         = {Dual-Layer Semantic-Spatial Belief Mapping for Aerial Object Goal Navigation},
  author        = {Xiao, Jianqiang and Deng, Xiang and Sun, Yuexuan and Wu, Yanjin and Yan, Wenbiao and Nie, Liqiang},
  journal={arXiv preprint \href{https://arxiv.org/abs/2609.08164}{arXiv:2609.08164}},
  year          = {2026},
}

@inproceedings{chen2023semanticfrontiers,
  author    = {Chen, Junting and Li, Guohao and Kumar, Suryansh
               and Ghanem, Bernard and Yu, Fisher},
  title     = {How To Not Train Your Dragon: Training-Free Embodied Object Goal Navigation with Semantic Frontiers},
  booktitle = {Proceedings of Robotics: Science and Systems (RSS)},
  year      = {2023},
}

@inproceedings{zhou2023esc,
  author    = {Zhou, Kaiwen and Zheng, Kaizhi and Pryor, Connor and Shen, Yilin and Jin, Hongxia and Getoor, Lise and Wang, Xin Eric},
  title     = {{ESC}: Exploration with Soft Commonsense Constraints for Zero-Shot Object Navigation},
  booktitle = {Proceedings of the 40th International Conference on Machine Learning (ICML)},
  volume    = {202},
  pages     = {42829--42842},
  year      = {2023},
}

@inproceedings{blip,
  title={Blip: Bootstrapping language-image pre-training for unified vision-language understanding and generation},
  author={Li, Junnan and Li, Dongxu and Xiong, Caiming and Hoi, Steven},
  booktitle={Proceedings of the 39th International Conference on Machine Learning (ICML)},
  volume = {162},
  pages={12888--12900},
  year={2022},
}

@inproceedings{Gd,
  title={Grounding dino: Marrying dino with grounded pre-training for open-set object detection},
  author={Liu, Shilong and Zeng, Zhaoyang and Ren, Tianhe and Li, Feng and Zhang, Hao and Yang, Jie and Jiang, Qing and Li, Chunyuan and Yang, Jianwei and Su, Hang and others},
  booktitle={European Conference on Computer Vision (ECCV)},
  pages={38--55},
  year={2024},
}

@article{HM3D,
  title={Habitat-matterport 3d dataset (hm3d): 1000 large-scale 3d environments for embodied ai},
  author={Ramakrishnan, Santhosh K and Gokaslan, Aaron and Wijmans, Erik and Maksymets, Oleksandr and Clegg, Alex and Turner, John and Undersander, Eric and Galuba, Wojciech and Westbury, Andrew and Chang, Angel X and others},
  journal={arXiv preprint \href{https://arxiv.org/abs/2109.08238}{arXiv:2109.08238}},
  year={2021}
}

@article{MP3D,
  title={Matterport3d: Learning from rgb-d data in indoor environments},
  author={Chang, Angel and Dai, Angela and Funkhouser, Thomas and Halber, Maciej and Niessner, Matthias and Savva, Manolis and Song, Shuran and Zeng, Andy and Zhang, Yinda},
  journal={arXiv preprint \href{https://arxiv.org/abs/1709.06158}{arXiv:1709.06158}},
  year={2017}
}

@article{Zson,
  title={Zson: Zero-shot object-goal navigation using multimodal goal embeddings},
  author={Majumdar, Arjun and Aggarwal, Gunjan and Devnani, Bhavika and Hoffman, Judy and Batra, Dhruv},
  journal={Advances in Neural Information Processing Systems (NeurIPS)},
  volume={35},
  pages={32340--32352},
  year={2022}
}

@inproceedings{L3mvn,
  title={L3mvn: Leveraging large language models for visual target navigation},
  author={Yu, Bangguo and Kasaei, Hamidreza and Cao, Ming},
  booktitle={2023 IEEE/RSJ International Conference on Intelligent Robots and Systems (IROS)},
  pages={3554--3560},
  year={2023},
}

@inproceedings{PSL,
  title={Prioritized semantic learning for zero-shot instance navigation},
  author={Sun, Xinyu and Liu, Lizhao and Zhi, Hongyan and Qiu, Ronghe and Liang, Junwei},
  booktitle={European Conference on Computer Vision (ECCV)},
  pages={161--178},
  year={2024},
}

@article{Stairway,
  title={Stairway to Success: An Online Floor-Aware Zero-Shot Object-Goal Navigation Framework via LLM-Driven Coarse-to-Fine Exploration},
  author={Gong, Zeying and Li, Rong and Hu, Tianshuai and Qiu, Ronghe and Kong, Lingdong and Zhang, Lingfeng and Zhao, Guoyang and Ding, Yiyi and Liang, Junwei},
  journal={IEEE Robotics and Automation Letters (RAL)},
  year={2026},
}

@article{A,
  author  = {Hart, Peter E. and Nilsson, Nils J. and Raphael, Bertram},
  title   = {A Formal Basis for the Heuristic Determination of Minimum Cost Paths},
  journal = {IEEE Transactions on Systems Science and Cybernetics},
  volume  = {4},
  number  = {2},
  pages   = {100--107},
  year    = {1968}
}

@article{mobile_sam,
  title   = {Faster Segment Anything: Towards Lightweight {SAM} for Mobile Applications},
  author  = {Zhang, Chaoning and Han, Dongshen and Qiao, Yu and
             Kim, Jung Uk and Bae, Sung-Ho and Lee, Seungkyu and
             Hong, Choong Seon},
  journal = {arXiv preprint arXiv:2306.14289},
  year    = {2023}
}

@inproceedings{busch2025onemap,
  author    = {Busch, Finn Lukas and Homberger, Timon and
               Ortega-Peimbert, Jes{\'u}s and Yang, Quantao and
               Andersson, Olov},
  title     = {One Map to Find Them All: Real-time Open-Vocabulary Mapping
               for Zero-shot Multi-Object Navigation},
  booktitle = {2025 IEEE International Conference on Robotics and Automation (ICRA)},
  year      = {2025},
  pages     = {14835--14842},
  doi       = {10.1109/ICRA55743.2025.11128393}
}
\bibliographystyle{iclr2027_conference}

\newpage
\appendix
\section{Appendix}
\label{sec:A}

\subsection{Spatial Support Construction}
\label{app:spatial_belief}

\paragraph{Camera-Aligned Coordinates.}
Let $\mathbf{p}_t$ denote the camera position on the
2D map. Let $\mathbf{e}_{r,t}$ and $\mathbf{e}_{a,t}$
be orthogonal unit vectors perpendicular and parallel
to the camera heading, respectively.
For a map location $\mathbf{q}$, we define
\begin{equation}
\begin{aligned}
r_t(\mathbf{q})
&=
\mathbf{e}_{r,t}^{\top}
(\mathbf{q}-\mathbf{p}_t),
\\
a_t(\mathbf{q})
&=
\mathbf{e}_{a,t}^{\top}
(\mathbf{q}-\mathbf{p}_t).
\end{aligned}
\label{eq:spatial_coordinates}
\end{equation}
Given the support center $\mathbf{q}_t^0$ obtained
from depth and camera pose, the two Gaussian components
are
\begin{equation}
\begin{aligned}
G_{r,t}(r)
&=
\mathcal{N}(r;\mu_{r,t},\sigma_{r,t}^{2}),
\\
G_{a,t}(a)
&=
\mathcal{N}(a;\mu_{a,t},\sigma_{a,t}^{2}),
\end{aligned}
\label{eq:spatial_gaussians}
\end{equation}
where
$\mu_{r,t}=r_t(\mathbf{q}_t^0)$ and
$\mu_{a,t}=a_t(\mathbf{q}_t^0)$.
The positive bandwidths $\sigma_{r,t}$ and
$\sigma_{a,t}$ determine the lateral and longitudinal
support extents.

\paragraph{Ray-Based Visibility Gating.}
To prevent support from extending behind occluders,
we cast a ray from $\mathbf{p}_t$ to $\mathbf{q}$
on the occupancy map.
Let $\mathcal{A}_t$ contain cells within the effective
field of view and sensing range that admit a valid
association with the current RGB-D observation.
Let $\mathcal{F}_t$ contain known-free cells after
incorporating the current depth observation.
The set $\mathcal{R}_t(\mathbf{q})$ contains the
intermediate ray cells, excluding the camera cell
and the endpoint cell.
The visibility gate is
\begin{equation}
\chi_t(\mathbf{q})
=
\mathbb{I}\!\left[
\mathbf{q}\in\mathcal{A}_t
\right]
\prod_{\mathbf{u}\in\mathcal{R}_t(\mathbf{q})}
\mathbb{I}\!\left[
\mathbf{u}\in\mathcal{F}_t
\right],
\label{eq:spatial_visibility}
\end{equation}
where $\mathbb{I}[\cdot]$ is the indicator function
and an empty product equals one.
An occupied or unknown intermediate cell blocks the
association, whereas an occupied endpoint can still
receive support.
This test is a planar approximation under the
2D occupancy representation.

\paragraph{Peak-Normalized Spatial Support.}
We combine the Gaussian components through their
peak-normalized product:
\begin{equation}
B_{\mathrm{sp},t}(\mathbf{q})
=
\chi_t(\mathbf{q})
\frac{
G_{r,t}(r_t(\mathbf{q}))
G_{a,t}(a_t(\mathbf{q}))
}{
G_{r,t}(\mu_{r,t})
G_{a,t}(\mu_{a,t})
}.
\label{eq:spatial_normalization}
\end{equation}
Expanding the Gaussian ratio gives
Eq.~\ref{eq:spatial_belief}.
Peak normalization bounds the support score in
$[0,1]$, while visibility gating suppresses invalid
or occluded associations.

\subsection{Action Transitions}
\label{app:action transitions}


\paragraph{Path Following and View Alignment.}
The action converter translates a selected waypoint or camera
view into discrete simulator actions.
We first compute an A* path through reachable, collision-free
space on the current occupancy map.
For the next local path target $\mathbf{g}$, the heading error is
$e_\theta =
\operatorname{wrap}
(\operatorname{atan2}(g_y-p_y,g_x-p_x)-\theta)$,
where $(\mathbf{p},\theta)$ is the current planar pose and
$\operatorname{wrap}$ maps angles to $(-\pi,\pi]$.
The controller issues \texttt{TURN\_LEFT} or
\texttt{TURN\_RIGHT} when $|e_\theta|>15^\circ$;
otherwise, it issues \texttt{MOVE\_FORWARD} if the predicted
motion is collision-free.
Each forward action advances $0.25\,\mathrm{m}$, and each
turn changes the heading by $30^\circ$.
The pose and local target are updated after each action,
and an invalidated path is replanned.

For an active-observation view
$v=(\mathbf{p}_v,\theta_v)$, the controller first reaches
$\mathbf{p}_v$ within its positional tolerance and then
performs in-place turns to the nearest executable heading
to $\theta_v$.
Candidate visibility is evaluated at the corresponding
discretized terminal pose, rather than an unattainable
continuous pose.
Although the task action space includes
\texttt{LOOK\_UP} and \texttt{LOOK\_DOWN},
this planar converter controls translation and yaw
while leaving camera pitch unchanged.

\paragraph{Action Accounting and Feedback.}
Let $\boldsymbol{a}_t(v)=(a_1,\ldots,a_{N_v})$ denote the
planned action sequence for view $v$.
The active-observation cost is $c_t(v)=N_v$, including
forward actions, intermediate turns, and final heading
alignment.
Only views satisfying $N_v\leq C_{\max}$ are eligible,
with $C_{\max}=8$ by default.
All executed actions count toward both the refinement
budget and the episode's interaction budget; replanning
does not reset the remaining refinement budget.

\subsection{LLMs Prompts}
\label{app:prompt}

\paragraph{Target-Conditioned description generation.}
The semantic belief uses $M$ descriptions of the same
target category.
The descriptions vary visual attributes and
associated-object cues without changing the target.
The resulting prompt set is kept fixed across observations
of that category.
For the prompt-count analysis, smaller sets are prefixes
of the same ordered description list.

\begin{quote}
\small
\textbf{Input:} Target category \texttt{TARGET};
required number of descriptions \texttt{M}.

\textbf{Instruction:}
Generate \texttt{M} concise visual descriptions for
image--text matching of \texttt{TARGET}.
Each description must explicitly refer to the same target
category.
Vary its visual description and plausible associated-object
context, while keeping the target central.
Do not replace the target with a related object or treat
contextual cues as proof that the target is present.
Do not assume any particular scene or target location.
Return only an ordered list of descriptions.
\end{quote}

\paragraph{Inputs for Semantic Reassessment.}
Within a triggered neighborhood, currently visible cells
with sufficient spatial support and high semantic
uncertainty are mapped back to RGB pixels using depth
and camera pose.
Connected pixel regions form padded crops, with the
selected evidence highlighted.
Each crop is supplied with the target category and its
associated BLIP-2 prompt--score pairs.
Historical inputs are restricted to observations
associated with the same map region and target category.

\paragraph{Shared Reliability Instruction.}
Both assessment stages evaluate the reliability of the
existing prompt-conditioned semantic assessment, not
the probability that the target is present.

\begin{quote}
\small
\textbf{Instruction:}
Assess whether the supplied visual evidence supports
the current BLIP-2 target-relevance assessment.
The similarity scores are not calibrated probabilities.
Distinguish direct target evidence, contextual cues,
and visually similar non-target objects.
Use an integer reliability score from 1 to 10:
1 indicates that the assessment is unsupported or
contradicted, and 10 indicates that it is clearly supported.
A high score does not necessarily imply target presence.
For example, a clearly non-target region can receive
high reliability when its existing assessment also
indicates low target relevance.
Provide a brief evidence summary rather than a
replacement similarity score.
\end{quote}

\paragraph{Intra-Observation Assessment.}

\begin{quote}
\small
\textbf{Input:}
The target category, current highlighted crop,
and associated BLIP-2 prompt--score pairs.

\textbf{Instruction:}
Evaluate whether the current semantic assessment is
justified by the highlighted region.
Use the surrounding crop only as context.
Identify visible target-specific features, related-object
cues, and plausible confounders.
Check whether the score pattern is supported by the image
or is driven mainly by contextual associations.
Account for occlusion, truncation, and insufficient detail
when judging reliability.
Return \texttt{r\_intra}, an integer from 1 to 10,
and a one-sentence evidence summary.
\end{quote}

\paragraph{Inter-Observation Assessment.}

\begin{quote}
\small
\textbf{Input:}
The current crop, prompt--score pairs, and
intra-observation assessment, together with available
historical crops and their associated scores for the
same map region and target category.

\textbf{Instruction:}
First determine whether the historical observations
provide comparable evidence for the highlighted region.
Then assess whether they support or contradict the
current semantic assessment.
Distinguish genuine contradictions from changes caused
by viewpoint, occlusion, or incomplete image coverage.
An object not being visible in another view is not,
by itself, evidence of contradiction.
Return \texttt{comparable}, \texttt{r\_inter},
and a one-sentence evidence summary.
If no comparable historical evidence is available,
set \texttt{comparable} to false and
\texttt{r\_inter} to null.
Otherwise, return an integer \texttt{r\_inter}
from 1 to 10 using the shared reliability criterion.
\end{quote}

\paragraph{Use of the Outputs.}
For region $j$, the scores determine
$\gamma_{t,j}
=\min(r_{t,j}^{\mathrm{intra}},
r_{t,j}^{\mathrm{inter}})/10$;
without comparable history,
$\gamma_{t,j}=r_{t,j}^{\mathrm{intra}}/10$.
The factor changes the semantic variance to
$\sigma_{\mathrm{sem},t}^{2}/\gamma_{t,j}$
while leaving $\mu_{\mathrm{sem},t}$ unchanged.
The original prompt statistics remain available for
uncertainty assessment.
Reassessment replaces the semantic contribution of the
corresponding observation rather than adding a new,
independent measurement.

\subsection{Rules for Generating Historical Waypoints}
\label{app:rules for waypoint}

After each movement step, the agent evaluates whether to
store its current position $\mathbf{p}_t$ as a historical
waypoint, considering both observation coverage and the
step interval since the last waypoint insertion.

Let $\mathcal{D}_t$ denote the map cells within the current
camera field of view and sensing range before occlusion
filtering. Applying the visibility gate $\chi_t$ to the
full current RGB-D observation, we measure spatial coverage as
\begin{equation}
C_{\mathrm{view},t}
=
\frac{\sum_{\mathbf{q}\in\mathcal{D}_t}\chi_t(\mathbf{q})}
{\max(1,|\mathcal{D}_t|)}.
\label{eq:historical_view_coverage}
\end{equation}
This score measures the fraction of the nominal viewing
footprint with valid, unoccluded observation support.

Let $\Delta_t$ be the number of navigation steps since
the most recent historical waypoint was stored.
We define the admission score as
\begin{equation}
S_{\mathrm{hist},t}
=
C_{\mathrm{view},t}
\frac{\Delta_t}{\Delta_t+T_{\mathrm{hist}}},
\qquad T_{\mathrm{hist}}>0,
\label{eq:historical_admission_score}
\end{equation}
where $T_{\mathrm{hist}}$ controls how quickly the
step-dependent weight increases.
The current position is stored when
$S_{\mathrm{hist},t}\geq\tau_{\mathrm{hist}}$,
with $\tau_{\mathrm{hist}}\in(0,1)$.
Thus, broader observation coverage favors admission,
while a longer step interval increases the admission
score for the same coverage.

The step counter resets after each insertion; before
the first insertion, it counts from the episode start.
Stored positions form the waypoint memory, whose
currently reachable entries constitute
$\mathcal{V}_t^{\mathrm{hist}}$ for joint evaluation
with frontier waypoints.

\subsection{Baseline Description} 
\label{app:baseline description}
We compare RefBeliefNav with nine zero-shot ObjectNav baselines. ZSON~\citep{Zson} transfers image-goal policies to text-specified goals. ESC~\citep{zhou2023esc} guides exploration through commonsense-based soft constraints, while L3MVN~\citep{L3mvn} uses language-model priors for frontier selection. PSL~\citep{PSL} improves semantic generalization through prioritized image-goal training. VLFM~\citep{yokoyama2024vlfm} guides frontier exploration using a vision-language value map. OpenFMNav~\citep{kuang2024openfmnav} supports free-form instructions through semantic score mapping. SG-Nav~\citep{yin2024sgnav} combines hierarchical scene graphs with LLM reasoning. BeliefMapNav~\citep{zhou2025beliefmapnav} integrates semantic and spatial evidence into a 3D target-belief map. ASCENT~\citep{Stairway} addresses multi-floor navigation through floor-aware mapping and coarse-to-fine reasoning.
\subsection{Additional Ablation Studies.}
\label{sec:Parameter Sensitivity Analyses}

\paragraph{Effectiveness of LLMs and VLMs.} As shown in Table~\ref{tab:ablation3}, GPT-4o achieves the
highest SR (69.5\%) and SPL (35.5\%), followed by Gemini
with 69.1\% SR and 35.3\% SPL, and Qwen2.5-VL with
68.9\% SR and 34.9\% SPL. The relatively small differences
suggest that RefBeliefNav maintains comparable performance
across the evaluated LLM backbones. Table~\ref{tab:ablation4} compares different VLM backbones.
BLIP-2~\cite{Blip-2} achieves the highest SR (69.5\%) and
SPL (35.5\%), slightly outperforming CLIP~\cite{Clip}
at 69.3\% SR and 35.2\% SPL. Both outperform BLIP,
which obtains 68.2\% SR and 35.1\% SPL.

\begin{table}[htbp]
    \centering
    
    \begin{minipage}{0.48\textwidth}
        \centering
        \caption{Impact of the LLMs.}
        \label{tab:ablation3}
        \setlength{\tabcolsep}{5pt}
        \begin{tabular}{ccc}
        \toprule
        Model & SR$\uparrow$ & SPL$\uparrow$ \\
        \midrule
        Gemini       & 69.1 & 35.3 \\
        Qwen2.5-VL & 68.9 & 34.9 \\
        \textbf{GPT-4o}  & \textbf{69.5} & \textbf{35.5} \\
        \bottomrule
        \end{tabular}
    \end{minipage}\hfill 
    \begin{minipage}{0.48\textwidth}
        \centering
        \caption{Impact of the VLMs.}
        \label{tab:ablation4}
        \setlength{\tabcolsep}{5pt}
        \begin{tabular}{ccc}
        \toprule
        Model & SR$\uparrow$ & SPL$\uparrow$ \\
        \midrule
        Blip\cite{blip}       & 68.2 & 35.1 \\
        Clip\cite{Clip} & 69.3 & 35.2 \\
        \textbf{Blip-2}\cite{Blip-2}  & \textbf{69.5} & \textbf{35.5} \\
        \bottomrule
        \end{tabular}
    \end{minipage}
    
\end{table}

\paragraph{Effectiveness of the Trigger Mechanism.}
As shown in Table ~\ref{tab:trigger_ablation}, adaptive triggering achieves the highest
SR (69.5\%) and SPL (35.5\%) with only 8.2 LLM calls.
In comparison, triggering every 10 steps yields 66.7\% SR
and 33.5\% SPL with 37.2 calls, while triggering every
30 steps achieves 66.1\% SR and 33.2\% SPL with 13.2 calls.
These results suggest that refinement timing matters more
than frequency alone: selectively reassessing ambiguous
evidence can improve navigation success and efficiency
while reducing LLM usage.

\begin{table}[t]
    \centering
    \small
    \setlength{\tabcolsep}{7pt}
    \caption{Impact of refinement-triggering
    strategy.}
    \label{tab:trigger_ablation}
    \begin{tabular}{cccc}
        \toprule
        Trigger Strategy & LLM Calls$\downarrow$& SR$\uparrow$ & SPL$\uparrow$ \\
        \midrule
        Fixed steps (per 10)
        &37.2 & 66.7 & 33.5 \\
        Fixed steps (per 30)
        &13.2 & 66.1 & 33.2 \\
        \textbf{Adaptive}
        & \textbf{8.2} & \textbf{69.5} & \textbf{35.5} \\
        \bottomrule
    \end{tabular}
\end{table}

\paragraph{Observation Window and Semantic Statistics.}
Table~\ref{tab:sensitivity_semantic} varies the observation-window
length $K$, prompt count $M$, and semantic standard-deviation
floor $\varepsilon$.
Changing $K$ affects recent-evidence uncertainty estimation,
but does not reset the recursively accumulated target belief
or delete archived observations.
The prompt-count comparison uses nested prompt sets so that
larger sets retain the prompts from smaller sets.
The variance-floor comparison examines the balance between
excessive observation precision and overly weak semantic
updates.
\begin{table}[t]
    \centering

    \begin{minipage}[t]{0.48\linewidth}
        \vspace{0pt}
        \centering
        \caption{Sensitivity analysis of the observation
        window and semantic statistics on HM3D.}
        \label{tab:sensitivity_semantic}

        \footnotesize
        \setlength{\tabcolsep}{3pt}
        \renewcommand{\arraystretch}{0.95}

        \begin{tabular*}{\linewidth}{
            @{\extracolsep{\fill}}lccc@{}
        }
            \toprule
            Parameter & Value & SR$\uparrow$ & SPL$\uparrow$ \\
            \midrule
            $K$ & 1  & 66.3 & 33.4 \\
                & 3  & 68.9 & 34.7 \\
                & \textbf{5}
                & \textbf{69.5} & \textbf{35.5} \\
                & 10 & 68.9 & 34.5 \\
                & 20 & 69.1 & 34.9 \\
            \midrule
            $M$ & 2  & 66.3 & 33.6 \\
                & 4  & 68.3 & 34.8 \\
                & \textbf{8}
                & \textbf{69.5} & \textbf{35.5} \\
                & 16 & 69.4 & 35.2 \\
            \midrule
            $\varepsilon$
                & 0.002 & 68.1 & 34.4 \\
                & 0.006 & 68.8 & 35.1 \\
                & \textbf{0.020}
                & \textbf{69.5} & \textbf{35.5} \\
                & 0.060 & 68.7 & 34.9 \\
                & 0.200 & 65.3 & 32.6 \\
            \bottomrule
        \end{tabular*}
    \end{minipage}\hfill
    \begin{minipage}[t]{0.48\linewidth}
        \vspace{0pt}
        \centering
        \caption{Sensitivity analysis of spatial support
        bandwidths on HM3D.}
        \label{tab:sensitivity_spatial}

        \footnotesize
        \setlength{\tabcolsep}{3pt}
        \renewcommand{\arraystretch}{0.95}

        \begin{tabular*}{\linewidth}{
            @{\extracolsep{\fill}}lccc@{}
        }
            \toprule
            Parameter & Value & SR$\uparrow$ & SPL$\uparrow$ \\
            \midrule
            $\sigma_{r,t}$
                & 0.250 & 66.8 & 34.0 \\
                & 0.375 & 68.5 & 35.0 \\
                & \textbf{0.500}
                & \textbf{69.5} & \textbf{35.5} \\
                & 0.750 & 69.3 & 35.1 \\
                & 1.000 & 68.6 & 34.4 \\
            \midrule
            $\sigma_{a,t}$
                & 0.50 & 66.5 & 33.8 \\
                & 0.75 & 68.6 & 35.0 \\
                & \textbf{1.00}
                & \textbf{69.5} & \textbf{35.5} \\
                & 1.50 & 69.3 & 35.0 \\
                & 2.00 & 68.1 & 34.1 \\
            \bottomrule
        \end{tabular*}
    \end{minipage}
\end{table}

\paragraph{Spatial Gaussian Bandwidths.}
Table~\ref{tab:sensitivity_spatial} varies the radial and
axial support bandwidths independently.
These parameters control the spatial extent of evidence
association.
Smaller bandwidths concentrate support around the projected
center, whereas larger bandwidths spread the same observation
over a broader map region.

\paragraph{Trigger Neighborhood and Belief Thresholds.}
Table~\ref{tab:sensitivity_trigger} varies the local radius
$R$, belief-gap threshold $\delta$, and relevance threshold
$\tau_{\mathrm{rel}}$.
The radius determines which nearby waypoints participate
in the trigger decision.
The gap threshold controls when competing waypoint beliefs
are considered similar, while the relevance threshold
prevents refinement of candidates with insufficient target
relevance.
These comparisons examine the balance between missing
potentially useful refinement opportunities and activating
refinement too broadly.

\paragraph{Active-Observation Geometry and Action Budget.}
Table~\ref{tab:sensitivity_ao} examines the admissible
observation distances, angular coverage, and action budget.
The distance bounds are restricted to the valid sensing
range.
The angular comparison changes the visibility criterion
through $\phi_{\mathrm{obs}}/\phi_{\mathrm{cam}}$ without
altering the actual camera field of view.
The action-budget comparison varies the maximum number
of simulator actions available to one active-observation
operation, while retaining the same overall navigation
budget.

\begin{table}[t]
    \centering

    \begin{minipage}[t]{0.48\linewidth}
        \vspace{0pt}
        \centering
        \caption{Sensitivity analysis of trigger
        neighborhood and belief thresholds.}
        \label{tab:sensitivity_trigger}

        \footnotesize
        \setlength{\tabcolsep}{3pt}
        \renewcommand{\arraystretch}{0.95}

        \begin{tabular*}{\linewidth}{
            @{\extracolsep{\fill}}lccc@{}
        }
            \toprule
            Parameter & Value & SR$\uparrow$ & SPL$\uparrow$ \\
            \midrule
            $R$ (m)
                & 1.50 & 66.7 & 34.0 \\
                & 2.25 & 68.7 & 35.1 \\
                & \textbf{3.00}
                & \textbf{69.5} & \textbf{35.5} \\
                & 4.50 & 69.4 & 34.8 \\
                & 6.00 & 68.6 & 33.6 \\
            \midrule
            $\delta$
                & 0.01 & 67.2 & 34.3 \\
                & 0.02 & 68.6 & 35.1 \\
                & \textbf{0.03}
                & \textbf{69.5} & \textbf{35.5} \\
                & 0.05 & 69.3 & 35.0 \\
                & 0.08 & 68.5 & 34.1 \\
            \midrule
            $\tau_{\mathrm{rel}}$
                & 0.15 & 68.8 & 34.1 \\
                & 0.20 & 69.3 & 35.0 \\
                & \textbf{0.25}
                & \textbf{69.5} & \textbf{35.5} \\
                & 0.30 & 68.2 & 34.9 \\
                & 0.35 & 66.4 & 33.8 \\
            \bottomrule
        \end{tabular*}
    \end{minipage}\hfill
    \begin{minipage}[t]{0.48\linewidth}
        \vspace{0pt}
        \centering
        \caption{Sensitivity analysis of active-observation
        geometry and action budget.}
        \label{tab:sensitivity_ao}

        \footnotesize
        \setlength{\tabcolsep}{3pt}
        \renewcommand{\arraystretch}{0.95}

        \begin{tabular*}{\linewidth}{
            @{\extracolsep{\fill}}lccc@{}
        }
            \toprule
            Parameter & Value & SR$\uparrow$ & SPL$\uparrow$ \\
            \midrule
            $d_{\min}$ (m)
                & 0.25 & 68.4 & 34.6 \\
                & \textbf{0.50}
                & \textbf{69.5} & \textbf{35.5} \\
                & 0.75 & 68.8 & 35.0 \\
                & 1.00 & 67.5 & 34.2 \\
            \midrule
            $d_{\max}$ (m)
                & 2.0 & 67.2 & 34.3 \\
                & \textbf{3.0}
                & \textbf{69.5} & \textbf{35.5} \\
                & 4.0 & 69.3 & 35.0 \\
                & 5.0 & 68.8 & 34.5 \\
            \midrule
            $\phi_{\mathrm{obs}}/\phi_{\mathrm{cam}}$
                & 0.50 & 66.9 & 33.7 \\
                & 0.75 & 68.6 & 34.8 \\
                & \textbf{1.00}
                & \textbf{69.5} & \textbf{35.5} \\
            \midrule
            $C_{\max}$ (actions)
                & 4  & 66.8 & 34.2 \\
                & 6  & 68.7 & 35.1 \\
                & \textbf{8}
                & \textbf{69.5} & \textbf{35.5} \\
                & 12 & 69.4 & 34.7 \\
                & 16 & 69.3 & 33.6 \\
            \bottomrule
        \end{tabular*}
    \end{minipage}
\end{table}

\paragraph{Planning Weights and Distance Decay.}
Table~\ref{tab:sensitivity_planning} varies the uncertainty
weight $\lambda_{\mathrm{unc}}$, navigation-cost weight
$\lambda_{\mathrm{cost}}$, and distance-decay scale $\ell$.
The zero-weight configurations disable the corresponding
planning terms examined in Table~\ref{tab:planning_ablation}.
These comparisons assess the contributions of uncertainty-driven
exploration and navigation-cost regularization to search
success and path efficiency.

\paragraph{Spatial and Semantic Uncertainty Thresholds.}
Table~\ref{tab:sensitivity_uncertainty} varies
$\tau_{\mathrm{sp}}$ and $\tau_{\mathrm{sem}}$.
Lower thresholds make the corresponding refinement branch
eligible more frequently, whereas higher thresholds
restrict refinement to stronger evidence deficiencies.
Because $\tau_{\mathrm{sem}}$ also appears in semantic
uncertainty normalization for search planning, its sweep
changes both refinement triggering and planning scores.

\begin{table}[t]
    \centering

    \begin{minipage}[t]{0.48\linewidth}
        \vspace{0pt}
        \centering
        \caption{Sensitivity analysis of planning weights
        and distance decay.}
        \label{tab:sensitivity_planning}

        \footnotesize
        \setlength{\tabcolsep}{3pt}
        \renewcommand{\arraystretch}{0.95}

        \begin{tabular*}{\linewidth}{
            @{\extracolsep{\fill}}lccc@{}
        }
            \toprule
            Parameter & Value & SR$\uparrow$ & SPL$\uparrow$ \\
            \midrule
            $\lambda_{\mathrm{unc}}$
                & 0.000 & 58.2 & 33.8 \\
                & 0.125 & 68.4 & 35.0 \\
                & \textbf{0.250}
                & \textbf{69.5} & \textbf{35.5} \\
                & 0.500 & 68.8 & 34.2 \\
                & 1.000 & 66.7 & 32.6 \\
            \midrule
            $\lambda_{\mathrm{cost}}$ ($\mathrm{m}^{-1}$)
                & 0.0 & 66.7 & 30.4 \\
                & 2.5 & 69.3 & 34.8 \\
                & \textbf{5.0}
                & \textbf{69.5} & \textbf{35.5} \\
                & 10.0 & 67.8 & 35.1 \\
                & 20.0 & 64.8 & 33.9 \\
            \midrule
            $\ell$ (m)
                & 1.0 & 66.5 & 34.2 \\
                & 2.0 & 68.6 & 35.1 \\
                & \textbf{3.0}
                & \textbf{69.5} & \textbf{35.5} \\
                & 4.5 & 69.0 & 34.8 \\
                & 6.0 & 68.3 & 34.1 \\
            \bottomrule
        \end{tabular*}
    \end{minipage}\hfill
    \begin{minipage}[t]{0.48\linewidth}
        \vspace{0pt}
        \centering
        \caption{Sensitivity analysis of uncertainty
        thresholds.}
        \label{tab:sensitivity_uncertainty}

        \footnotesize
        \setlength{\tabcolsep}{3pt}
        \renewcommand{\arraystretch}{0.95}

        \begin{tabular*}{\linewidth}{
            @{\extracolsep{\fill}}lccc@{}
        }
            \toprule
            Parameter & Value & SR$\uparrow$ & SPL$\uparrow$ \\
            \midrule
            $\tau_{\mathrm{sp}}$
                & 0.30 & 69.4 & 34.1 \\
                & 0.40 & 69.3 & 35.0 \\
                & \textbf{0.50}
                & \textbf{69.5} & \textbf{35.5} \\
                & 0.60 & 68.4 & 34.9 \\
                & 0.70 & 66.8 & 34.0 \\
            \midrule
            $\tau_{\mathrm{sem}}$
                & 0.0025 & 68.6 & 33.9 \\
                & 0.0050 & 69.1 & 34.9 \\
                & \textbf{0.0100}
                & \textbf{69.5} & \textbf{35.5} \\
                & 0.0200 & 68.2 & 34.5 \\
                & 0.0400 & 66.5 & 33.5 \\
            \bottomrule
        \end{tabular*}
    \end{minipage}
\end{table}

\paragraph{Joint Parameter Sensitivity.}
Table~\ref{tab:sensitivity_joint} supplements the one-factor
comparisons with two local parameter grids.
The first varies $\lambda_{\mathrm{unc}}$ jointly with
$\lambda_{\mathrm{cost}}$ to examine the interaction between
uncertainty-driven exploration and travel efficiency.
The second varies $\delta$ jointly with $\tau_{\mathrm{sp}}$
to examine the interaction between waypoint-belief ambiguity
and spatial evidence deficiency.
Entries along the reference row or column match the
corresponding one-factor tables.

\begin{table}[t]
    \centering
    \caption{Joint parameter sensitivity:
    (a) planning weights and (b) trigger thresholds.
    $\lambda_{\mathrm{cost}}$ is expressed in
    $\mathrm{m}^{-1}$.}
    \label{tab:sensitivity_joint}

    \begin{minipage}[t]{0.48\linewidth}
        \vspace{0pt}
        \centering
        \footnotesize
        \setlength{\tabcolsep}{3pt}
        \renewcommand{\arraystretch}{0.95}

        \begin{tabular*}{\linewidth}{
            @{\extracolsep{\fill}}cccc@{}
        }
            \toprule
            \multicolumn{4}{c}{\textbf{(a) Planning weights}} \\
            \midrule
            $\lambda_{\mathrm{unc}}$
            & $\lambda_{\mathrm{cost}}$
            & SR$\uparrow$
            & SPL$\uparrow$ \\
            \midrule
            0.125 & 2.5  & 68.7 & 34.7 \\
            0.125 & 5.0  & 68.4 & 35.0 \\
            0.125 & 10.0 & 67.0 & 34.5 \\
            0.250 & 2.5  & 69.3 & 34.8 \\
            \textbf{0.250} & \textbf{5.0}
            & \textbf{69.5} & \textbf{35.5} \\
            0.250 & 10.0 & 67.8 & 35.1 \\
            0.500 & 2.5  & 69.1 & 33.8 \\
            0.500 & 5.0  & 68.8 & 34.2 \\
            0.500 & 10.0 & 68.0 & 34.5 \\
            \bottomrule
        \end{tabular*}
    \end{minipage}\hfill
    \begin{minipage}[t]{0.48\linewidth}
        \vspace{0pt}
        \centering
        \footnotesize
        \setlength{\tabcolsep}{3pt}
        \renewcommand{\arraystretch}{0.95}

        \begin{tabular*}{\linewidth}{
            @{\extracolsep{\fill}}cccc@{}
        }
            \toprule
            \multicolumn{4}{c}{\textbf{(b) Trigger thresholds}} \\
            \midrule
            $\delta$
            & $\tau_{\mathrm{sp}}$
            & SR$\uparrow$
            & SPL$\uparrow$ \\
            \midrule
            0.02 & 0.40 & 68.9 & 34.9 \\
            0.02 & 0.50 & 68.6 & 35.1 \\
            0.02 & 0.60 & 67.9 & 34.7 \\
            0.03 & 0.40 & 69.3 & 35.0 \\
            \textbf{0.03} & \textbf{0.50}
            & \textbf{69.5} & \textbf{35.5} \\
            0.03 & 0.60 & 68.4 & 34.9 \\
            0.05 & 0.40 & 69.4 & 34.6 \\
            0.05 & 0.50 & 69.3 & 35.0 \\
            0.05 & 0.60 & 68.7 & 34.8 \\
            \bottomrule
        \end{tabular*}
    \end{minipage}
\end{table}

\end{document}